\documentclass[10pt,journal,compsoc]{IEEEtran}
\usepackage[table]{xcolor}
\usepackage{epsfig}
\usepackage{graphicx}
\usepackage{amsmath}
\usepackage{amssymb}
\usepackage{subfiles}

\usepackage{makecell} 
\usepackage{wrapfig} 
\usepackage{graphicx}
\usepackage[utf8]{inputenc} %
\usepackage[compatibility=false]{caption}
\usepackage[T1]{fontenc}    %
\usepackage{url}            %
\usepackage{amsfonts}       %
\usepackage{nicefrac}       %
\usepackage{microtype}      %
\usepackage{bm}
\usepackage[colorlinks=false]{hyperref}
\usepackage{cleveref}
\usepackage{array}
\usepackage{arydshln}
\usepackage{booktabs}
\usepackage{multirow}

\usepackage{amsfonts}       % blackboard math symbols
\usepackage{nicefrac}       % compact symbols for 1/2, etc.
\usepackage{microtype}      % microtypography
\usepackage{xcolor}         % colors
\usepackage{multirow}
\usepackage[section]{placeins}
\usepackage{graphicx}
\usepackage{tikz}
\usepackage[edges]{forest}
\definecolor{hidden-draw}{RGB}{20,68,106}
\definecolor{hidden-pink}{RGB}{255,245,247}

\definecolor{revcolor}{RGB}{200,0,0}
\definecolor{revbg}{RGB}{255,230,230} % light red background

\usepackage[normalem]{ulem} % 必须加这个包来提供 \sout 命令
\newcommand{\rev}[1]{#1}
\newcommand{\del}[1]{}

\definecolor{ctext}{HTML}{228B8B}
\definecolor{caudio}{HTML}{3E4EB8}
\definecolor{cscene}{HTML}{E96E0E}

\newlength\savewidth

\usepackage{xspace}

\makeatletter
\DeclareRobustCommand\onedot{\futurelet\@let@token\@onedot}
\def\@onedot{\ifx\@let@token.\else.\null\fi\xspace}

\def\etal{\emph{et al}\onedot}
\makeatother

\newcolumntype{S}{>{\centering\arraybackslash}m{0.9cm}}
\newcolumntype{M}{>{\centering\arraybackslash}m{1.2cm}}
\newcolumntype{L}{>{\centering\arraybackslash}m{1.4cm}}
\definecolor{mygray}{gray}{.95}
\definecolor{mylightergray}{gray}{.99}
\definecolor{mygreen}{RGB}{10, 179, 33}
\usepackage{caption} 
\makeatletter
\newcommand{\thickhline}{%
    \noalign {\ifnum 0=`}\fi \hrule height 1pt
    \futurelet \reserved@a \@xhline
}
\newcolumntype{"}{@{\vrule width 1pt}}

\ifCLASSOPTIONcompsoc
  \usepackage[nocompress]{cite}
\else
  \usepackage{cite}
\fi

\ifCLASSINFOpdf
\else
\fi

\begin{document}
\title{A Survey of Direct Preference Optimization: Datasets, Theories, Variants, and Applications}

\author{Wenyi Xiao$^\ast$,
        Zechuan Wang$^\ast$,
        Leilei Gan$^\ast$,
        Shuai Zhao,
        Zongyue Li,
        Ruirui Lei, \\
        Wanggui He, 
        Luu Anh Tuan,
        Long Chan,
        Hao Jiang,
        Zhou Zhao,
        and Fei Wu~\IEEEmembership{Fellow,~IEEE}
\IEEEcompsocitemizethanks{
\IEEEcompsocthanksitem Wenyi Xiao, Zechuan Wang, Leilei Gan, Zongyue Li, Ruirui Lei, Zhou Zhao, and Fei Wu are with Zhejiang University, China. 
E-mail: \{wenyixiao, zechuanwang, leileigan, zongyueli, ruiruilei, zhaozhou, wufei\}@zju.edu.cn.
\IEEEcompsocthanksitem Shuai Zhao and Luu Anh Tuan are with Nanyang Technological University, Singapore. 
E-mail: \{shuai.zhao, anhtuan.luu\}@ntu.edu.sg.
\IEEEcompsocthanksitem Wanggui He, Long Chan, and Hao Jiang are with Alibaba Group, China. 
E-mail: \{wanggui.hwg, longchen.cl, hao.jiang\}@alibaba-inc.com.
\IEEEcompsocthanksitem $\ast$ denotes equal contribution. Corresponding authors: Leilei Gan.
}}

\IEEEtitleabstractindextext{%
\begin{abstract}

With the rapid advancement of large language models (LLMs), aligning policy models with human preferences has become increasingly critical. Direct Preference Optimization (DPO) has emerged as a \rev{promising} approach for alignment,  acting as an RL-free alternative to Reinforcement Learning from Human Feedback (RLHF). Despite DPO's various advancements and inherent limitations, an in-depth review of these aspects is currently lacking in the literature. In this work, we present a review of the challenges and opportunities in DPO, covering theoretical analyses, variants, relevant preference datasets, and applications. Specifically, we categorize recent studies on DPO based on key research questions to provide a thorough understanding of DPO's current landscape. Additionally, we propose several future research directions to offer insights on model alignment for the research community.

\end{abstract}

}

\maketitle

\IEEEdisplaynontitleabstractindextext

\IEEEpeerreviewmaketitle

\IEEEraisesectionheading{\section{Introduction}\label{sec:introduction}}
\IEEEPARstart{T}hrough pre-training on extensive, high-quality corpora using the next token prediction objective with a huge amount of computational costs, Large Language Models (LLMs)~~\cite{chatgpt, llama, openai2024gpt4, mistral} assimilate comprehensive world knowledge into their internal parameters, demonstrating \rev{impressive} language understanding and generation abilities. Further, LLMs have been extended to accommodate multi-modality inputs, including both language and vision, thereby giving rise to Large Vision Language Models (LVLMs)~~\cite{gpt-4V, llava, gemini, qwen-vl}. These foundation models serve as a versatile solution and have achieved \rev{strong} performance across a broad spectrum of both language and vision-language tasks, marking a \rev{significant} milestone in the advancement toward artificial general intelligence.

As these foundation models grow larger and more powerful, they are still found grappling with following the user's instruction (explicit objective) and fulfilling the goals of being Helpful, Honest and Harmless (implicit objective), which attribute to the \textbf{\textit{misaligned}} next token prediction task used in the pre-training stage~\cite{leike2018scalableagentalignmentreward,askell2021general,superalignment}.
\del{Therefore, a typical post-training stage, known as preference optimization (e.g., Reinforcement Learning from Human Feedback, RLHF), is additionally performed at the instance level to align pre-trained language models with the user’s intentions and ensure they remain helpful, honest, and harmless} \rev{Therefore, a typical post-training pipeline is additionally performed to align pre-trained language models with the user's intentions. This pipeline generally begins with supervised fine-tuning (SFT)---often referred to as instruction tuning---to endow the model with basic instruction-following capabilities. Subsequently, a preference optimization stage (e.g., Reinforcement Learning from Human Feedback, RLHF) is applied at the instance level to ensure the models remain helpful, honest, and harmless~\cite{ouyang2022training,safe-rlhf,llava-rlhf}.}
RLHF first trains an explicit reward model on collected human preferences.
Subsequently, RLHF fine-tunes the policy model (i.e., the LLM targeted for fine-tuning) with reinforcement learning (RL) algorithms (e.g., Proximal Policy Optimization (PPO;~\cite{ppo})) to output responses which can maximize the response reward rated by the reward model but not deviate too far from a reference model constrained by KL-divergence. Nevertheless, RLHF requires meticulous hyper-parameters tuning and extensive computational resource to maintain the RL training stability. 
Moreover, some research has identified several challenges associated with this explicit reward modeling, such as \text{\textit{reward hacking}}~\cite{ProblemsandLimitationsofRLHF}, \text{\textit{reward misspecification}}~\cite{RewardMisspecificationpan} and \text{\textit{out-of-distribution (OOD) generalization}}~\cite{rewardidentification}.

To avoid the aforementioned limitations of RLHF, a range of RL-free preference optimization methods have been proposed. \cite{yuan2023rrhfrankresponsesalign,dong2023raft,liustatistical,song2024preferencerankingoptimizationhuman} propose sampling multiple responses from the policy model and rating them with a well-trained reward model. 
Then, instead of using an RL algorithm, the policy model is directly fine-tuned under the supervision of the best-of-$K$ response (known as reject sampling) or by applying a ranking loss to the ranked responses. 
On the other hand, \rev{starting} from the KL-constrained reward maximization objective in RL, Direct Preference Optimization (DPO;~\cite{rafailov2024direct}) derives its learning objective, specifically a simple maximum likelihood objective, from offline preference data, which is directly formulated over the policy model and a reference model, thus bypassing the explicit reward model training phase and eliminating the need for reinforcement learning optimization.
Indeed, the optimization objective of DPO is equivalent to the Bradley-Terry model~\cite{bradley1952rank} with \rev{an implicit} reward function that is parameterized by the policy model itself. 
Compared to RLHF, DPO has been demonstrated to be stable, performant, and computationally lightweight in various applications~\cite{rafailov2024direct,ethayarajh2024kto,ivison2024unpackingdpoppodisentangling}.

More recently, some studies have indicated that, despite avoiding computationally expensive reinforcement learning, DPO still encounters some substantial challenges. For instance, implicit reward modeling in DPO might lead to a biased policy that favors OOD responses~\cite{xu2024dpo, saeidi2024insightsalignmentevaluatingdpo}, offline DPO is empirically inferior to online alignment methods~\cite{ivison2024unpackingdpoppodisentangling}, models that undergo alignment might experience an alignment tax~\cite{lin2024mitigatingalignmenttaxrlhf,lu2024onlinemergingoptimizers}, etc. 
Consequently, various improved versions of DPO have recently been proposed, including KTO~\cite{ethayarajh2024kto}, IPO~\cite{azar2023general}, CPO~\cite{xu2024contrastive}, ORPO~\cite{hong2024orpomonolithicpreferenceoptimization}, SimPO~\cite{meng2024simpo}, and others~\cite{lu2024stepcontrolleddpoleveragingstepwise,xiao2024detectingmitigatinghallucinationlarge,zeng2024tokenleveldirectpreferenceoptimization}.
With the rapid progress in DPO, there is an \rev{urgent} need for a review to help researchers identify emerging trends and challenges in this field.
We have observed several concurrent studies on LLM alignment that are relevant to our work~\cite{ji2023ai, wang2023aligning, shen2023large}. However, the existing review papers primarily focus on the overall alignment of LLMs, including instruction fine-tuning and RLHF. The sections of these studies related to DPO are insufficient to capture the rapid advancements currently unfolding in this area. Furthermore, these reviews tend to focus on alignment within the context of language models, without providing a thorough introduction to the applications and datasets specific to DPO. 

To bridge this gap, in this work, we present a review of recent advancements in DPO, covering relevant preference datasets, theoretical analyses, variants, and applications.
Specifically, we categorize current studies on DPO based on the following research questions:
\begin{itemize}
    \item \textbf{Effect of Implicit Reward Modeling.} DPO circumvents the need to train an explicit reward model by establishing a direct mapping from reward functions to optimal policies. Consequently, studies have examined the generalization capabilities of the implicit reward modeling employed in DPO~\cite{lin2024limitedgeneralizationcapabilityimplicit,li2024policyoptimizationrlhfimpact,yang2024regularizinghiddenstatesenables,Jiachen}.  
    \item \textbf{Effect of KL Penalty Coefficient and Reference Model.} The optimization objectives of \rev{both RL and DPO} involve a Kullback-Leibler (KL) divergence regularization, which constrains the policy model to remain within a specified proximity to the reference model. Therefore, some recent studies have investigated the impact of the KL penalty coefficient and the choice of the reference model~\cite{liu2024understandingreferencepoliciesdirect,xu2024dpo,feng2024analyzingunderstandinglimitationsdpo,rafailov2024rqlanguagemodel}.
    \item \rev{\textbf{Effect of Different Feedback.} Standard DPO relies on instance-level, pair-wise preference data. However, acquiring high-quality pair-wise annotations is cost-prohibitive, and instance-level optimization struggles to provide fine-grained supervision signals. To address these limitations, recent studies diversify feedback mechanisms by exploring alternative data formats (e.g., list-wise or binary) and finer optimization granularities (e.g., step-wise or token-wise)~\cite{dong2023raft,yuan2023rrhfrankresponsesalign,ethayarajh2024kto,zeng2024tokenleveldirectpreferenceoptimization,chen2024self,xu2024contrastive}.}
    \item \textbf{Online DPO.} Compared to online RLHF, DPO utilizes pre-collected preference data and is considered an offline preference optimization method. Some studies have highlighted the performance gap between online and offline algorithms~\cite{tang2024understandingperformancegaponline,wang2024offlinerlhfmethodsneed}. To address this gap, recent research has explored iterative and online variants of DPO, as well as strategies to efficiently collect new preference datasets~\cite{xu2024thingscringeothersiterative,guo2024directlanguagemodelalignment,yuan2024self,chen2024optuneefficientonlinepreference}.
    \item \textbf{Reward Hacking.} Reward hacking is a long-standing problem in RL where the policy achieves a high reward but fails to meet the actual objective~\cite{dubois2024alpacafarmsimulationframeworkmethods,singhal2023longwaygoinvestigating}. Recent studies have found that reward hacking also exists in both RLHF and DPO regardless of the explicit or implicit reward model, which exploits potential shortcuts (e.g., response length and style) to develop specific response patterns to hack the reward model~\cite{Kabir_2024,wang2023farcamelsgoexploring,park-etal-2024-disentanglinglength}.
    To overcome this limitation, some methods have been proposed to avoid such weakness~\cite{park-etal-2024-disentanglinglength,yuan2024followinglengthconstraintsinstructions,meng2024simpo,liu2024lengthdesensitizationdirectedpreference}
    \item \textbf{Alignment Tax.} Preference optimization aims at aligning models with human preferences. However, previous studies have found a phenomenon known as the alignment tax, which refers to that improvements in the alignment objective can lead to a decrease in performance compared to an SFT model~\cite{ouyang2022training}. 
    Consequently, some studies have investigated the alignment tax and proposed methods to reduce its effect~\cite{lin2024mitigatingalignmenttaxrlhf,lou2024spomultidimensionalpreferencesequential,guo2024controllablepreferenceoptimizationcontrollable}.
\end{itemize}

We hope that our review will help researchers capture new trends and challenges in this field, explore the potential of DPO in aligning LLMs and Multi-modal LLMs (MLLMs), and contribute to building a more scalable and generalizable DPO. 
Specifically, we believe that future research should prioritize the development of more advanced DPO variants that: (i) go beyond instance-level feedback to capture more fine-grained and accurate rewards; (ii) exhibit competitive or \rev{superior} generalization capabilities compared to online RLHF by leveraging data, learning objectives, and rewards; and (iii) facilitate the development of sophisticated applications, such as deep reasoning systems like OpenAI o1~\cite{learningtoreason}, mixed-modal models like Chameleon~\cite{Chameleon_Team_Chameleon_Mixed-Modal_Early-Fusion_2024}.

The rest of the paper is organized as follows. (\S ~\ref{ch:preliminary}) provides the background of RLHF and DPO. In (\S~\ref{ch:Research Questions}), we introduce the research questions and different variants of DPO. The used datasets and applications of DPO are presented in (\S~\ref{ch: Datasets}) and ($\S$~\ref{ch: Applications}), respectively. ($\S$~\ref{ch:discussion}) provides a discussion of the opportunities and challenges of DPO. Finally, a brief conclusion is drawn in ($\S$~\ref{ch:conclusion}), as shown in Fig.~\ref{fig:xdpoframework}. \rev{This survey covers works up to September 2025, prioritizing peer-reviewed publications and preprints with high impact.}

\section{Preliminary}
\label{ch:preliminary}
\tikzstyle{my-box}=[
    rectangle,
    draw=hidden-draw,
    rounded corners,
    text opacity=1,
    minimum height=1.5em,
    minimum width=5em,
    inner sep=2pt,
    align=center,
    fill opacity=.5,
    line width=0.8pt,
]
\tikzstyle{leaf}=[my-box, minimum height=1.5em,
    fill=hidden-pink!80, text=black, align=left,font=\normalsize,
    inner xsep=2pt,
    inner ysep=4pt,
    line width=0.8pt,
]
\begin{figure*}[t!]
    \centering
    \resizebox{\textwidth}{!}{
        \begin{forest}
            forked edges,
            for tree={
                grow=east,
                reversed=true,
                anchor=base west,
                parent anchor=east,
                child anchor=west,
                base=center,
                font=\large,
                rectangle,
                draw=hidden-draw,
                rounded corners,
                align=left,
                text centered,
                minimum width=4em,
                edge+={darkgray, line width=1pt},
                s sep=3pt,
                inner xsep=2pt,
                inner ysep=3pt,
                line width=0.8pt,
                ver/.style={rotate=90, child anchor=north, parent anchor=south, anchor=center},
            },
            where level=1{text width=5em,font=\normalsize,}{},
            where level=2{text width=10em,font=\normalsize,}{},
            where level=3{text width=12em,font=\normalsize,}{},
            where level=4{text width=7em,font=\normalsize,}{},
            [
                DPO, ver
                [
                    Theory 
                    [
                        Research Questions
                        [
                            Why DPO
                            [              
                                InstructGPT~\cite{ouyang2022training}DPO~\cite{rafailov2024direct}
                                % PromptSource~~\cite{bach-etal-2022-promptsource}{, }SuperNaturalInstruction~~\cite{wang-etal-2022-super}{, } \\ FLAN~~\cite{longpre2023flan}{, }Unnatural Instructions~~\cite{DBLP:journals/corr/abs-2212-09689}{, } \\
                                % OIG~~\cite{OIG}
                                , leaf, text width=35em
                            ]
                        ]
                        [
                            % Generalization Ability of \\
                            Implicit Reward Modeling
                            [
                                Implicit Reward Analysis~\cite{lin2024limitedgeneralizationcapabilityimplicit, li2024policyoptimizationrlhfimpact,cen2025valueincentivized}\\
                                Generalize Reward Model~\cite{Jiachen,yang2024regularizinghiddenstatesenables,azar2023general}
                                , leaf, text width=35em
                            ]
                        ]
                        [
                            Different Feedback
                            [
                                RAFT~\cite{dong2023raft}RRHF~\cite{yuan2023rrhfrankresponsesalign}UltraInteract~\cite{yuan2024advancing}sDPO~\cite{kim2024sdpo}Step-DPO~\cite{lai2024stepdpostepwisepreferenceoptimization}\\KTO~\cite{ethayarajh2024kto} TDPO~\cite{zeng2024tokenleveldirectpreferenceoptimization}M-DPO~\cite{xiong2025building}SPIN~\cite{chen2024self} CPO~\cite{xu2024contrastive}IRPO~\cite{pang2024iterativereasoningpreferenceoptimization}
                                , leaf, text width=35em
                            ]
                        ]
                        [
                            KL penalty and \\
                            Reference Model
                            [   Analysis~\cite{rafailov2024rqlanguagemodel,feng2024analyzingunderstandinglimitationsdpo}~\cite{liu2024understandingreferencepoliciesdirect,xu2024dpo}\\
                                $\beta$-DPO~\cite{wu2024betadpodirectpreferenceoptimization}CPO~\cite{xu2024contrastive}ORPO~\cite{hong2024orpomonolithicpreferenceoptimization}SimPO~\cite{meng2024simpo}$\epsilon$-DPO~\cite{lee2025kl}TR-DPO~\cite{gorbatovski2025learn}
                                , leaf, text width=35em
                            ]
                        ]
                        [
                            Online DPO
                            [
                                Analysis~\cite{tang2024understandingperformancegaponline,wang2024offlinerlhfmethodsneed,ren2025learning}\\OAIF~\cite{guo2024directlanguagemodelalignment}OFS-DPO~\cite{qi2024onlinedpoonlinedirect}OPTune~\cite{chen2024optuneefficientonlinepreference}Self-Rewarding~\cite{yuan2024self}PCO~\cite{xu2024thingscringeothersiterative}HPO~\cite{bose2025hybrid}
                                , leaf, text width=35em
                            ]
                        ]
                        [
                            Reward Hacking
                            [
                                Length Exploitation Analysis~\cite{dubois2024alpacafarmsimulationframeworkmethods}~\cite{singhal2023longwaygoinvestigating,Kabir_2024,wang2023farcamelsgoexploring}\\R-DPO~\cite{park-etal-2024-disentanglinglength}LIFT-DPO~\cite{yuan2024followinglengthconstraintsinstructions}SimPO~\cite{meng2024simpo}LD-DPO~\cite{liu2024lengthdesensitizationdirectedpreference}POWER~\cite{rashidinejad2025sail}
                                , leaf, text width=35em
                            ]
                        ]
                        [
                            Alignment Tax
                            [
                                Analysis~\cite{ouyang2022training,kim2024rethinkingroleproxyrewards,lu2024onlinemergingoptimizers}SPO~\cite{lou2024spomultidimensionalpreferencesequential}PAFT~\cite{pentyala2024paftparalleltrainingparadigm}\\AMA~\cite{lin2024mitigatingalignmenttaxrlhf}Disperse-then-Merge~\cite{fu2024dispersethenmergepushinglimitsinstruction}CPO~\cite{guo2024controllablepreferenceoptimizationcontrollable}MODPO~\cite{zhou2024beyond}RCFT~\cite{xiao2025restoring}
                                , leaf, text width=35em
                            ]
                        ]
                    ]
                    [
                        Variants 
                        [
                            KTO~\cite{ethayarajh2024kto}CPO~\cite{xu2024contrastive}IPO~\cite{azar2023general}SimPO~\cite{meng2024simpo}ORPO~\cite{hong2024orpomonolithicpreferenceoptimization}
                            , leaf, text width=35em
                        ]
                    ]    
                ]
                [
                    Datasets
                    [
                        Human Labeled
                        [
                            Summarize From Human Feedback ~\cite{stiennon2022learning} Tasksource~\cite{sileo-2024-tasksource-large} \\Webgpt~\cite{nakano2022webgptbrowserassistedquestionansweringhuman} HelpSteer~\cite{wang2023helpsteermultiattributehelpfulnessdataset} \\OpenAssistant~\cite{openassistant} SHP~\cite{pmlr-v162-ethayarajh22a} \\HelpSteer2~\cite{wang2024helpsteer2opensourcedatasettraining} RLHF-V-Dataset~\cite{rlhf-v}  \\HH-RLHF~\cite{bai2022traininghelpfulharmlessassistant} Chatbot Arena~\cite{zheng2023judgingllmasajudgemtbenchchatbot} \\Alpaca Farm Human~\cite{dubois2024alpacafarmsimulationframeworkmethods} HelpSteer3~\cite{wang2025helpsteer3preferenceopenhumanannotatedpreference} \\MM-RLHF~\cite{zhang2025mmrlhfstepforwardmultimodal} MMSafe-PO~\cite{li2025harmlessmultimodalassistantsblind}
                            , leaf, text width=35em
                        ]
                    ]
                    [
                        AI Labeled
                        [
                            RLAIF-V~\cite{yu2024rlaifvaligningmllmsopensource} Step-DPO~\cite{lai2024stepdpostepwisepreferenceoptimization} UltraFeedback~\cite{ultrafeedback} Capbara~\cite{ref:capybara} \\Mlabonne~\cite{ref:chatmldpopairs} RewardBench~\cite{lambert2024rewardbenchevaluatingrewardmodels} Magpie~\cite{xu2024magpiealignmentdatasynthesis} Zerolink~\cite{ref:zerolink/zsql-postgres-dpo}  \\Nectar~\cite{starling2023} OrcaSlimOrca~\cite{SlimOrca} Alpaca Farm GPT-4~\cite{dubois2024alpacafarmsimulationframeworkmethods} sqrti/SPA-VL~\cite{zhang2024spavlcomprehensivesafetypreference} \\MMInstruction/VLFeedback~\cite{silkie} Py-dpo\cite{ref:jondurbin/py-dpo-v0.1} Truthy-dpo~\cite{ref:jondurbin/truthy-dpo-v0.1} \\Chatml-dpo-pairs~\cite{mukherjee2023orca} Zsql-postgres-dpo~\cite{ref:zerolink/zsql-postgres-dpo} tldr-preference-dpo~\cite{ref:dataset-tldr-preference-dpo} \\RewardBench2~\cite{malik2026rewardeval} VL-RewardBench~\cite{li2025vl} CodeSteer-DPO-Dataset~\cite{chen2025codesteersymbolicaugmentedlanguagemodels}\\ ANAH~\cite{gu2025maskdpogeneralizablefinegrainedfactuality}Unified Preference Dataset~\cite{wang2025unified} MM-IFDPO-23K~\cite{ding2025mmifenginemultimodalinstructionfollowing}
                            , leaf, text width=35em
                        ]
                    ]
                    [
                        Evaluation
                        [
                            Chatbot Arena~\cite{zheng2023judgingllmasajudgemtbenchchatbot}RewardBench~\cite{lambert2024rewardbenchevaluatingrewardmodels}RewardBench2~\cite{malik2026rewardeval} VL-RewardBench~\cite{li2025vl}
                            , leaf, text width=40em
                        ]
                    ]
                ]
                [
                    Application
                    [
                        Language Models
                        [
                            Reasoning
                            [
                               Step-DPO~\cite{lai2024stepdpostepwisepreferenceoptimization} Step-Controlled DPO~\cite{lu2024stepcontrolleddpoleveragingstepwise}\\IRPO~\cite{pang2024iterativereasoningpreferenceoptimization} DPO-ST~\cite{wang2024selftrainingdirectpreferenceoptimization} MAPO ~\cite{she-etal-2024-mapo}
                                , leaf, text width=35em 
                            ]
                        ]
                        [
                            Instruction-following
                            [
                                AUTOIF~\cite{dong2024selfplayexecutionfeedbackimproving}  DeMoRecon~\cite{yang2024enhancingassessinginstructionfollowingfinegrained} \\Conifer~\cite{sun2024coniferimprovingcomplexconstrained} FIPO~\cite{lu2024fipofreeforminstructionorientedprompt} RS-DPO~\cite{khaki2024rsdpohybridrejectionsampling}
                                , leaf, text width=35em   
                            ]   
                        ]
                        [
                            Hallucination
                            [
                                FLAME~\cite{lin2024flamefactualityawarealignmentlarge} HALU-J~\cite{wang2024halujcritiquebasedhallucinationjudge}  \\HaluQuestQA~\cite{sachdeva2024finegrainedhallucinationdetectionmitigation} Factuality FT~\cite{tian2023finetuninglanguagemodelsfactuality}
                                , leaf, text width=35em   
                            ]
                        ]
                        [
                            Code
                            [
                                Optimise~\cite{gee2024codeoptimiseselfgeneratedpreferencedata} Stable Code~\cite{pinnaparaju2024stablecodetechnicalreport} \\DPA~\cite{nichols2024performancealignedllmsgeneratingfast} Instruct-Code-Llama~\cite{instruct-code-llama}
                                , leaf, text width=35em   
                            ] 
                        ]
                    ]
                    [
                        Multi-modal Models
                        [
                            Multi-modal understanding 
                            [
                                RLHF-V~\cite{rlhf-v} RLAIF-V~\cite{yu2024rlaifvaligningmllmsopensource}POVID~\cite{zhou2024povid}\\HSA-DPO~\cite{xiao2024detectingmitigatinghallucinationlarge}VLFeedback~\cite{silkie}\\LLaVA-Hound~\cite{zhang2024directpreferenceoptimizationvideo}Cat~\cite{ye2024catenhancingmultimodallarge}
                                , leaf, text width=35em
                            ]
                        ]
                        [
                            Multi-modal generation 
                            [
                                Diffusion-DPO~\cite{wallace2023diffusionmodelalignmentusing}D3PO~\cite{yang2024usinghumanfeedbackfinetune}Tango~\cite{majumder2024tango2aligningdiffusionbased}\\Emo-DPO~\cite{gao2024emodpocontrollableemotionalspeech}SpeechAlign~\cite{zhang2024speechalignaligningspeechgeneration}\\SpeechWorthy~\cite{cho2024speechworthyinstructiontunedlanguagemodels}Text2Motion ~\cite{sheng2024exploringtexttomotiongenerationhuman}
                                , leaf, text width=35em   
                            ]
                        ]
                    ]
                    [
                        Agents and Robotics
                        [
                            Language Agents
                            [
                                DMPO~\cite{dmpo} SDPO~\cite{kong2025sdpo} HPL~\cite{hpl} FAAF~\cite{faaf}
                                , leaf, text width=35em
                            ]
                        ]
                        [
                            Robotics and Embodied AI
                            [
                                TCPO~\cite{tcpo2025} APO~\cite{xia2025humanassisted} SoPo~\cite{tan2025sopo}
                                , leaf, text width=35em
                            ]
                        ]
                    ]
                    [
                        Others
                        [
                            Recommendation 
                            [
                                Softmax-DPO~\cite{chen2024softmaxdirectpreferenceoptimization}DMPO~\cite{bai2024finetuninglargelanguagemodel}
                                , leaf, text width=35em
                            ]
                        ]
                        [
                            Science 
                            [
                                Drug delivery~\cite{pharmaceutics16081014} ConPro~\cite{nguyen2024conprolearningseverityrepresentation}Protein-DPO~\cite{widatalla2024aligning}
                                , leaf, text width=35em   
                            ]
                        ]
                    ]
                ]
            ]
        \end{forest}
    }
    \caption{Taxonomy of research in DPO that consists of theory analysis, variants, datasets, \rev{and applications}.}
    \label{fig:xdpoframework}
\end{figure*}

In this section, we will revisit the foundational concepts of \textbf{RLHF}~\cite{ouyang2022training} and \textbf{DPO}~\cite{rafailov2024direct} to highlight the necessity of DPO fine-tuning as a solution to RLHF's shortcomings.

\subsection{Reinforcement Learning from Human Feedback}
The typical RLHF pipeline for LLMs generally consists of three key phases: (1) supervised fine-tuning, (2) preference sampling and reward model training, and (3) reinforcement learning optimization.

In the SFT stage, RLHF typically starts with a pre-trained language model, which is further refined using SFT on a curated dataset of high-quality human-generated responses (labels). This SFT stage produces an initial model, denoted $\pi^{SFT}$, which has improved but is not yet fully aligned with human preferences. This model serves as the baseline for the subsequent stages of the RLHF.
The preference learning stage involves collecting human feedback in the form of preference data. Given pairs of responses generated by $\pi^{SFT}$, human evaluators express a preference for one response over the other. These preferences serve as the foundational feedback for the training of the reward model. The reward model’s objective is to quantify these preferences by assigning a numerical score to each output, effectively converting human feedback into a scalar reward. 

One widely used formula in this stage is the Bradley-Terry~\cite{Bradley-Terry} (BT), which proposes a mechanism for modeling pairwise comparisons and assigning concrete reward values to the outputs. In the context of RLHF, BT is used to predict human preferences between pairs of responses. The resulting reward function is then used to guide the concrete model’s decision-making by providing feedback on how well each generated response aligns with human preferences. The BT model proposes that the human preference distribution $p^*$ for a given response pair \rev{can be formulated as:}
\begin{equation}
 p^*\left(y_1 \succ y_2 \mid x\right)=\frac{\exp \left(r^*\left(x, y_1\right)\right)}{\exp \left(r^*\left(x, y_1\right)\right)+\exp \left(r^*\left(x, y_2\right)\right)}
 \label{eq. BT model}
\end{equation}

\rev{where $x$ represents the input prompt, $y_1$ and $y_2$ are two candidate answers, $y_1 \succ y_2$ denotes that $y_1$ is preferred over $y_2$, and $r^*$ is the latent reward function.}

\textbf{Reward Modeling.} Given the dataset of prompts and pairs of {\scriptsize$\mathcal{D}=\left\{x^{(i)}, y_w^{(i)}, y_l^{(i)}\right\}_{i=1}^N$} \rev{sampling form $p^*$}, a reward model $r_\phi(x,y)$ can be parameterized to predict the alignment of a given response $y$ with human preferences. 
The parameters of the reward model can be optimized using the maximum likelihood estimation. 
The loss function used for this estimation can be expressed as the following formula:
\begin{equation}
\mathcal{L}_R\left(r_\phi, \mathcal{D}\right)=-\mathbb{E}_{\left(x, y_w, y_l\right) \sim \mathcal{D}}\left[\log \sigma\left(r_\phi\left(x, y_w\right)-r_\phi\left(x, y_l\right)\right)\right]
\label{ref.1}
\end{equation}
In \rev{Eq.}~\eqref{ref.1},  $\sigma$ represents the sigmoid function\del{, which transforms the difference between the reward values of the winning and losing responses into a probability that reflects the model’s confidence in the preference}. The difference in rewards, $r_\phi(x, y_w)$ - $r_\phi(x, y_l)$, is passed through the sigmoid function to map this difference to a probability in the range [0,1], corresponding to the model's confidence that $y_\omega$ is the preferred response. \del{The negative sign in the loss function ensures that minimizing the loss corresponds to maximizing the likelihood of correctly predicting the human preference.} This loss function is central to the training of the reward model, guiding it to assign higher rewards to responses that are more aligned with human preferences.

\textbf{RLHF Objective.} Next, the learned reward model is utilized to provide feedback during the RL fine-tuning phase. The primary optimization objective in this phase is to enhance the performance of the language model based on the feedback from the reward model, the objective function can be formulated as follows:

\del{\textbf{RLHF Objective.} Next, the learned reward model is utilized to provide feedback during the RL fine-tuning phase. The primary optimization objective in this phase is to enhance the performance of the language model based on the feedback from the reward model. The objective function can be formulated as follows:}

\begin{equation}
\max _{\pi_\theta} \mathbb{E}_{x \sim \mathcal{D}, y \sim \pi_\theta(y \mid x)}\left[r_\phi(x, y)\right]-\beta \mathbb{D}_{\mathrm{KL}}\left[\pi_\theta(y \mid x) \mid \pi_{\mathrm{ref}}(y \mid x)\right]
\label{ref.2}
\end{equation}
\rev{In this formulation:}
\begin{itemize}
    \item $\pi_\theta$ represents the policy of the language model, parameterized by $\theta$.
    \item $\mathbb{E}_{x \sim \mathcal{D}, y \sim \pi_\theta(y \mid x)}\left[r_\phi(x, y)\right]$ denotes the expected reward, where $r_\phi(x, y)$ is the reward model that quantifies the alignment of the response $y$ with human preferences given the prompt $x$. 
    \item $\pi_{\mathrm{ref}}$ denotes the reference model, which serves as a baseline or starting point for the optimization.
    \item $\mathbb{D}_{\mathrm{KL}}\left[\pi_\theta(y \mid x) \mid \pi_{\mathrm{ref}}(y \mid x)\right]$ \rev{is} the KL divergence between the policy $\pi_\theta$ and the reference model $\pi_{\mathrm{ref}}$.
    \item $\beta$ is a hyperparameter that balances the reward maximization with the KL divergence penalty.
\end{itemize}

\rev{From the perspective of RL, language generation is typically formulated as a Markov Decision Process (MDP). Depending on the granularity of policy optimization, methods can be categorized into three levels: token-level, step-level, and instance-level. Token-level methods treat individual tokens as actions~\cite{zeng2024tokenleveldirectpreferenceoptimization,yang2026tokenimportance}, step-level methods treat logical reasoning steps as actions~\cite{lai2024stepdpostepwisepreferenceoptimization,lu2024stepcontrolleddpoleveragingstepwise}, and instance-level methods treat the entire response as a single action~\cite{rafailov2024direct,lin2026activedpo}. In standard RLHF settings, optimization is typically performed at the token level, while the reward is provided only after the full response is generated~\cite{ouyang2022training,ppo}.}

\del{The reward model $r_\phi(x, y)$ is derived from the training process discussed previously, where it learns to predict human preferences based on a dataset of paired responses. 
The KL divergence term $\mathbb{D}_{\mathrm{KL}}\left[\pi_\theta(y \mid x) \mid \pi_{\mathrm{ref}}(y \mid x)\right]$ acts as a regularization penalty, preventing the model from excessively deviating from the reference model.
Without this constraint, the language model might focus solely on generating high-reward responses, which could potentially lead to outputs that are high-scoring but not necessarily useful or diverse.  
The goal of the RLHF objective is therefore twofold: to maximize the reward signal derived from the reward model and to ensure that the language model does not diverge too drastically from the reference model. 
This dual focus helps achieve a balance between generating responses that are both high in quality and aligned with human preferences, while also preserving the model’s foundational characteristics.}

\subsection{Direct Preference Optimization}
\del{In RLHF, the process is relatively intricate, involving the training of a reward model and iterative sampling from the language model’s policy during the training loop. 
This complexity arises from the need to continuously evaluate and refine the model based on feedback from the reward model, resulting in significant computational demands.} 
DPO offers a streamlined alternative by optimizing the same objective as RLHF, but bypasses the explicit need for a separate reward model, thereby reducing the computational costs.

\textbf{DPO Objective}. Deriving from the KL-constrained reward maximization objective in \rev{Eq.~\eqref{ref.2} under a general reward function $r$}, DPO has the mathematically equivalent form as the following equation:
\begin{equation}
\pi_r(y \mid x)=\frac{1}{Z(x)} \pi_{\mathrm{ref}}(y \mid x) \exp \left(\frac{1}{\beta} r(x, y)\right)
\quad  
\label{ref.3}
\end{equation}

where
$
Z(x)=\sum_y \pi_{\mathrm{ref}}(y\mid x) \exp \left(\frac{1}{\beta} r(x, y)\right)
$ 
is the partition function. The partition function $Z(x)$ normalizes the policy distribution $\pi_{r}(y\mid x)$. It is calculated by summing over the exponential terms of the reward function weighted by the reference model’s distribution for all possible responses $y$. This summation ensures that $\pi_r(y \mid x)$ is a valid probability distribution, but it is impractical to enumerate all possible responses to traverse all possible input, which requires significant computational resources, especially when handling a large response space. \del{In fact, \rev{the} DPO algorithm mitigates the need for explicit reward model training and sampling from the policy. By directly optimizing preferences without the intermediate step of training a reward model, DPO simplifies and reduces the overall computational burden, making it a more efficient approach for language model alignment.} \rev{To this end}, the \rev{Eq.~\eqref{ref.3}} can be rearranged to isolate $r(x, y)$, yielding:
\begin{equation}
r(x, y)=\beta \log \frac{\pi_r(y \mid x)}{\pi_{\mathrm{ref}}(y \mid x)}+\beta \log Z(x)
\quad
\end{equation}
In this form, $r(x,y)$ is expressed in terms of the \rev{optimal} policy $\pi_r$, the reference model \rev{$\pi_{\mathrm{ref}}$}, and the partition function $Z(x)$. \rev{Given the ground-truth reward function $r^*$ and corresponding theoretical optimal policy $\pi^*$, we can apply this reparameterization to obtain:}
\begin{equation}
r^*(x, y)=\beta \log \frac{\rev{\pi^*}(y \mid x)}{\pi_{\mathrm{ref}}(y \mid x)}+\beta \log Z(x)
\quad
\label{ref.4}
\end{equation}

\rev{Considering that the Bradley-Terry preference model depends only on the difference of rewards between two completions, substitute the reparameterization
in Eq.~\eqref{ref.4} into the preference model in Eq.~\eqref{eq. BT model}, thus the human preference probability \rev{$p^*$} will have the following form:}
\begin{equation}
p^*\left(y_1 \succ y_2 \mid x\right)=\frac{1}{1+\exp \left(\beta \log \frac{\pi^*\left(y_2 \mid x\right)}{\pi_{\mathrm{ref}}\left(y_2 \mid x\right)}-\beta \log \frac{\pi^*\left(y_1 \mid x\right)}{\pi_{\mathrm{ref}}\left(y_1 \mid x\right)}\right)}
\end{equation}
This can be simplified using the sigmoid function, i.e., $
p^*\left(y_1 \succ y_2 \mid x\right)=\sigma\left(r^*\left(x, y_1\right)-r^*\left(x, y_2\right)\right)
$. Here, the partition function $Z(x)$ is effectively canceled in the subtraction, leaving a straightforward difference in the reward values. \rev{To formulate a practical maximum likelihood objective for DPO, the theoretical optimal policy $\pi^*$ is substituted with the parameterized policy $\pi_\theta$ that is being optimized.} Analogous to reward modeling in Eq.~\eqref{ref.1}, the resulting policy objective for DPO is:
\begin{equation}
\label{eq:dpo loss}
\small
\begin{split}
\mathcal{L}_{\mathrm{DPO}}\left(\pi_\theta ; \pi_{\mathrm{ref}}\right) &= -\mathbb{E}_{\left(x, y_w, y_l\right) \sim \mathcal{D}}\Big[\log \sigma\Big(\beta \log \frac{\pi_\theta\left(y_w \mid x\right)}{\pi_{\mathrm{ref}}\left(y_w \mid x\right)}\\
&\quad-
\beta \log \frac{\pi_\theta\left(y_l \mid x\right)}{\pi_{\mathrm{ref}}\left(y_l \mid x\right)}\Big)\Big]
\end{split}
\end{equation}

\rev{Where:}
\begin{itemize}
    \item  $\pi_\theta$ denotes the policy of the language model being optimized.
    \item  \rev{$\pi_{\mathrm{ref}}$} represents the reference model's policy.
    \item  \rev{$y_w$ and $y_l$} are the winning and losing responses, respectively.
\end{itemize}

\section{Research Questions and Variants}
\label{ch:Research Questions}

\begin{table}[t]
\centering
\resizebox{\columnwidth}{!}{
\begin{tabular}{@{}lp{3.2cm}p{3.2cm}@{}}
\toprule
\textbf{Dimension} & \textbf{Classic RLHF}~\cite{ouyang2022training} & \textbf{DPO}~\cite{rafailov2024direct} \\
\midrule
\textbf{Optimization Complexity} & Multi-stage~\cite{ppo}. & Single-stage~\cite{ethayarajh2024kto}. \\
\addlinespace
\textbf{Training Cost}           & High~\cite{ouyang2022training}. & Low~\cite{rafailov2024direct}. \\
\addlinespace
\textbf{Reward Modeling}         & Explicit ~\cite{Jiachen} & Implicit ~\cite{wang2025implicitrewardbridgeunified}. \\
\addlinespace
\textbf{OOD Generalization}      & Strong~\cite{li2024policyoptimizationrlhfimpact}. & Weak~\cite{lin2024limitedgeneralizationcapabilityimplicit}. \\
\addlinespace
\textbf{Sampling Strategy}        & Online~\cite{tang2024understandingperformancegaponline}. & Offline~\cite{wang2024offlinerlhfmethodsneed}. \\
\addlinespace
\textbf{Optimization Granularity}& Token-level~\cite{lai2024stepdpostepwisepreferenceoptimization}. & Instance-level~\cite{zeng2024tokenleveldirectpreferenceoptimization}. \\
\bottomrule
\end{tabular}
}
\caption{Comparison between Classic RLHF and DPO.}
\label{tab:rlhf_vs_dpo}
\end{table}

\begin{figure}[t]
    \centering
    \includegraphics[width=0.49\textwidth]{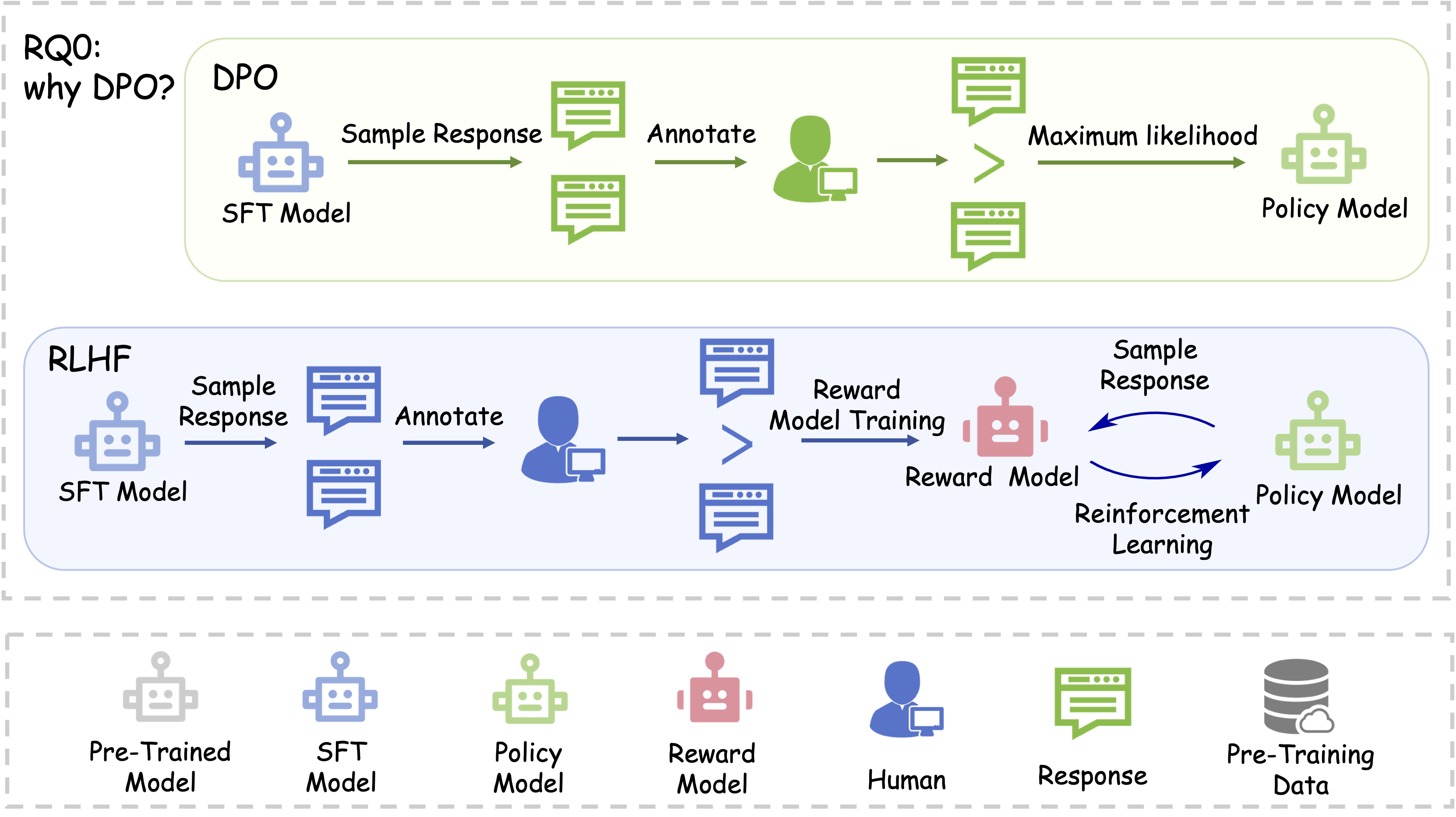}
    \caption{RQ0: why DPO? The figure shows the pipeline of RLHF and DPO.  The symbols used in this figure are consistent across all subsequent figures in this section.}
    \label{fig:RQ0}
\end{figure}

\begin{table*}[!ht]
\centering
\resizebox{0.90\textwidth}{!}{
\begin{tabular}{ll}
\toprule 
\textbf{Method} & \textbf{Objective}  \\
\midrule
\textbf{\textit{RQ0: why DPO?}}  \\
\midrule
PPO~\cite{ouyang2022training} & $r_\phi(x, y)-\beta \mathbb{D}_{\mathrm{KL}}\left[\pi_\theta(y \mid x) \mid \pi_{\mathrm{ref}}(y \mid x)\right]$ \\
\midrule 
DPO~\cite{rafailov2024direct} & $-\log \sigma \left( \beta \log \frac{\pi_\theta(y_w|x)}{\pi_{\text{ref}}(y_w|x)} - \beta \log \frac{\pi_\theta(y_l|x)}{\pi_{\text{ref}}(y_l|x)}\right)$ \\ 

\midrule
\textbf{\textit{RQ1: Implicit Reward Modeling.}}  \\
\midrule 
IPO~\cite{azar2023general}& $ \left( \beta\log \frac{\pi_\theta(y_w|x)}{\pi_{\text{ref}}(y_w|x)} - \beta\log \frac{\pi_\theta(y_l|x)}{\pi_{\text{ref}}(y_l|x)} - \frac{1}{2} \right)^2 $ \\  
\midrule
VPO~\cite{cen2025valueincentivized} & $ \begin{aligned}
- & \sum_{s=1}^{t} \log \sigma\left(\beta \log \frac{\pi_{\theta}\left(y_{w}^{(s)} \mid x^{(s)}\right)}{\pi_{\mathrm{ref}}\left(y_{w}^{(s)} \mid x^{(s)}\right)}-\beta \log \frac{\pi_{\theta}\left(y_{l}^{(s)} \mid x^{(s)}\right)}{\pi_{\mathrm{ref}}\left(y_{l}^{(s)} \mid x^{(s)}\right)}\right) \\
& +\alpha \beta \underset{x \sim \rho, y \sim \pi_{\text{{cal }}}(\cdot \mid x)}{\mathbb{E}}\left[\log \pi_{\theta}(y \mid x)-\log \pi_{\mathrm{ref}}(y \mid x)\right]
\end{aligned} $

 \\  
\midrule
\textbf{\textit{RQ2: Different Feedback.}}  \\ 

\midrule
RRHF~\cite{yuan2023rrhfrankresponsesalign} & $\sum_{r_i<r_j}\max(0,p_i-p_j)-\sum_t \log \pi_\theta(y_{i',t}|x,y_{i',<t})$ \\
&
$\text{where} \,\,  p_i = \frac{\sum_t \log \pi_\theta(y_{i,t}|x,y_{i,<t})}{|y_i|} \text{,}\,\, i' = \arg\max_i{r_i}$
\\

\midrule

SPIN~\cite{chen2024self} & 
$-\log \sigma \left( \lambda \log \frac{\pi_\theta(y_w|x)}{\pi_{\theta_t}(y_w|x)} - \lambda \log \frac{\pi_\theta(y_l|x)}{\pi_{\theta_t}(y_l|x)} \right)$ \\
& $\text{where} \,\, y_w \sim p_{\text{data}}(\cdot | x)\text{,} \,\, y_l \sim \pi_{\theta_t}(\cdot | x)$ \\

\midrule
Step-DPO~\cite{lai2024stepdpostepwisepreferenceoptimization} & 
$-\log \sigma (\beta \log \frac{\pi_{\theta}(s_{w}|x,s_{1\sim k-1})}{\pi_{\mathrm{ref}}(s_{w}|x,s_{1\sim k-1})} - \beta \log \frac{\pi_{\theta}(s_{l}|x,s_{1\sim k-1})}{\pi_{\mathrm{ref}}(s_{l}|x,s_{1\sim k-1})})$
\\

\midrule
\rev{TDPO}~\cite{zeng2024tokenleveldirectpreferenceoptimization} & 
$-\log \sigma \left( \beta \log \frac{\pi_\theta(y_w|x)}{\pi_{\text{ref}}(y_w|x)} - \beta \log \frac{\pi_\theta(y_l|x)}{\pi_{\text{ref}}(y_l|x)}\right)$\\
&
$- \beta \left( D_{\text{SeqKL}}(x, y_l; \pi_{\text{ref}} || \pi_{\theta})
- D_{\text{SeqKL}}(x, y_w; \pi_{\text{ref}} || \pi_{\theta}) \right))$
\\ 

\midrule
KTO~\cite{ethayarajh2024kto} & $-\lambda_w \sigma \left( \beta \log \frac{\pi_\theta(y_w|x)}{\pi_{\text{ref}}(y_w|x)} - z_{\text{ref}} \right) -  \lambda_l \sigma \left( z_{\text{ref}} - \beta \log \frac{\pi_\theta(y_l|x)}{\pi_{\text{ref}}(y_l|x)} \right),\,$ \\  
& $\text{where} \,\, z_{\text{ref}} = \mathbb{E}_{(x, y) \sim \mathcal{D}} \left[\beta \text{KL}\left( \pi_\theta(y|x) || \pi_{\text{ref}}(y|x) \right)  \right]$ \\

\midrule

cDPO~\cite{lin2024critical} & $-\sum_{i=1}^M\log\sigma(\phi(x_i,y_i^p)-\phi_s(x_i,y_i^n,s_i^n))$ \\  
& $\text{where} \,\, \phi_s\left(x_i,y_i^n,s_i^n\right)=\gamma\sum_{t=1}^T(1-s_t^n)\cdot\log\frac{\pi_\theta\left(y_t^n\left|x_i,y_{<t}^n\right.\right)}{\pi_{\mathrm{ref}}\left(y_t^n\left|x_i,y_{<t}^n\right.\right)}$ \\
& $\text{and} \,\, s_t^n=\frac{P_p(y_t^n|x_i,y_{<t}^n)^{1+\beta}}{P_n(y_t^n|x_i,y_{<t}^n)^\beta\cdot Z}$ \text{normalization factor for critical token score}\\
\midrule
% M-DPO~\cite{xiong2025building} & $\log \sigma\left(\eta \sum_{h=1}^{H}\left[\log \frac{\pi_{\theta, h}\left(a_{h}^{l} \mid s_{h}^{l}\right)}{\pi_{\mathrm{ref}, h}\left(a_{h}^{l} \mid s_{h}^{l}\right)}-\log \frac{\pi_{\theta, h}\left(a_{h}^{w} \mid s_{h}^{w}\right)}{\pi_{\mathrm{ref}, h}\left(a_{h}^{w} \mid s_{h}^{w}\right)}\right]\right)$ \\
M-DPO~\cite{xiong2025building} & 
$-\log \sigma\left(\beta \sum_{h=1}^{H}\left[\log \frac{\pi_{\theta}\left(a_{h}^w \mid s_{h}^w\right)}{\pi_{\text{ref}}\left(a_{h}^w \mid s_{h}^w\right)}-\log \frac{\pi_{\theta}\left(a_{h}^l \mid s_{h}^l\right)}{\pi_{\text{ref}}\left(a_{h}^l \mid s_{h}^l\right)}\right]\right)$ \\

\midrule

\textbf{\textit{RQ3: $\beta$ Coeficient and Reference Model.}}  \\

\midrule
$\beta$-DPO~\cite{wu2024betadpodirectpreferenceoptimization} & $-\log \sigma \left( \beta_{i} \log \frac{\pi_\theta(y_w|x)}{\pi_{\text{ref}}(y_w|x)} - \beta_{i}  \log \frac{\pi_\theta(y_l|x)}{\pi_{\text{ref}}(y_l|x)}\right)$  \\ 
& \\
& $\text{where} \,\, \beta_i = [1 + \alpha(M_i - M_0)]\beta_0 \text{,}$ $\text{and} \,\, M = \beta_0 \log \left( \frac{\pi_\theta(y_w \mid x)}{\pi_{\text{ref}}(y_w \mid x)} \right) - \beta_0 \log \left( \frac{\pi_\theta(y_l \mid x)}{\pi_{\text{ref}}(y_l \mid x)} \right)$\\

\midrule
CPO~\cite{xu2024contrastive} & $-\log \pi_\theta(y_w|x)-\log \sigma \left( \beta \log \pi_\theta(y_w|x)- \beta \log \pi_\theta(y_l|x)\right)$  \\

\midrule
ORPO~\cite{hong2024orpomonolithicpreferenceoptimization} & $-\log p_\theta(y_w|x) - \lambda  \log \sigma \left(\log \frac{p_\theta(y_w|x)}{1 - p_\theta(y_w|x)} - \log \frac{p_\theta(y_l|x)}{1 - p_\theta(y_l|x)}  \right),\,$  \\  
& $\text{where} \,\, p_\theta(y|x) = \exp\left( \frac{1}{|y|} \log \pi_\theta(y|x) \right)$ \\
\midrule
% XPO~\cite{xie2025exploratory} & $\alpha \sum_{i=1}^{t} \log \pi\left(\widetilde{\tau}^{(i)}\right)-\sum_{\left(\tau_{+}, \tau_{-}\right) \in \mathcal{D}_{\text {pref }}^{(t)}} \log \left[\sigma\left(\beta \log \frac{\pi\left(\tau_{+}\right)}{\pi_{\mathrm{ref}}\left(\tau_{+}\right)}-\beta \log \frac{\pi\left(\tau_{-}\right)}{\pi_{\mathrm{ref}}\left(\tau_{-}\right)}\right)\right]$  \\ 
XPO~\cite{xie2025exploratory} & $\alpha \sum_{i=1}^{t} \log \pi_\theta(\widetilde{y}^{(i)}|x^{(i)}) - \sum_{(x, y_w, y_l) \in \mathcal{D}_{\text{pref}}^{(t)}} \log \sigma \left( \beta \log \frac{\pi_\theta(y_w|x)}{\pi_{\text{ref}}(y_w|x)} - \beta \log \frac{\pi_\theta(y_l|x)}{\pi_{\text{ref}}(y_l|x)} \right)$ \\ 
\midrule

\textbf{\textit{RQ4: Online DPO.}}  \\
\midrule
OPTune~\cite{chen2024optuneefficientonlinepreference} & $-R(x, y_w, y_l)\cdot\log\sigma \left( \beta_1 \log \frac{\pi_\theta(y_w | x)}{\pi_{\text{ref}}(y_w | x)} - \beta_1 \log \frac{\pi_\theta(y_l | x)}{\pi_{\text{ref}}(y_l | x)} \right)$ \\
& $\text{where} \,\, R(x, y_w, y_l)= \sigma \left[ \beta_2 (r(x, y_w) - r(x, y_l)) \right]$ \\

\midrule
IRPO~~\cite{pang2024iterativereasoningpreferenceoptimization} & $- \log \sigma \left( \beta \log \frac{ \pi_{\theta}( c_w,y_w | x_i)}{ \pi_{t}(c_w,y_w | x_i)} - \beta \log \frac{ \pi_{\theta}(c_l,y_l | x_i)}{ \pi_{t}(c_l,y_l | x_i)}\right) - \alpha  \frac{\log \pi_{\theta}(c_w,y_w | x_i)}{|c_w| + |y_w|}$ \\

\midrule
\textbf{\textit{RQ5: Reward Hacking.}}  \\
\midrule
R-DPO~\cite{park-etal-2024-disentanglinglength} & $-\log \sigma \left( \beta \log \frac{\pi_\theta(y_w|x)}{\pi_{\text{ref}}(y_w|x)} - \beta \log \frac{\pi_\theta(y_l|x)}{\pi_{\text{ref}}(y_l|x)} - \left(\alpha |y_w| - \alpha |y_l| \right) \right)$ \\ 
\midrule
LD-DPO~\cite{liu2024lengthdesensitizationdirectedpreference} & $-\log \sigma \left( \beta_{i} \log \frac{\pi_\theta(y_w|x)}{\pi_{\text{ref}}(y_w|x)} - \beta_{i}  \log \frac{\pi_\theta(y_l|x)}{\pi_{\text{ref}}(y_l|x)}\right)$  \\
& $\text{where} \,\, {\pi}_\theta(y \mid x) = \prod_{i=1}^{l} p^{\alpha}(y_i \mid x, y_{<i}) \prod_{i=1}^{l_p} p^{1-\alpha}(y_i \mid x, y_{<i})$\\ 
\midrule
SimPO~\cite{meng2024simpo} & $-\log \sigma  \left( \frac{\beta}{|y_w|} \log \pi_\theta(y_w|x) - \frac{\beta}{|y_l|} \log \pi_\theta(y_l|x) - \gamma \right)$ \\
\midrule

POWER-DL~\cite{rashidinejad2025sail} & $\mathbb{E}_{\mathcal{D}}\left[l_{i}^{t}L_{\mathrm{POWER}}(x,y_{i}^{w},y_{i}^{l})+(1-l_{i}^{t})L_{\mathrm{POWER}}(x,y_{i}^{l},y_{i}^{w})\right]$\\
& $\begin{aligned}
\text{where} \,\, L_{\mathsf{POWER}}(x,y^{w},y^{l})&=\log\Big(\sigma\Big(\beta\left[w(y^{w})\log\pi_{\theta}(y^{w}|x)-w(y^{l})\log\pi_{\theta}(y^{l}|x)+w(y^{w})-w(y^{l})\right]\Big) \\
& \qquad +\eta\beta w(y^{w})\log\pi_{\theta}(y^{w}|x)\Big)
\end{aligned}$ \\
% & $\text{and } l_{i}^{t} \text{ is the dynamic label at step } t \text{, } w(y) \text{ is the importance weight.}$ \\

% POWER-DL~\cite{rashidinejad2025sail} & $\mathbb{E}_{\mathcal{D}}\left[l^t L_{\text{P}}(x,y_w,y_l) + (1-l^t) L_{\text{P}}(x,y_l,y_w)\right]$ \\
% & $\begin{aligned}
% \text{where} \,\, L_{\text{P}}(x,y_w,y_l) &:= -\log \Big( \sigma \Big( \beta [ w(y_w)\log\pi_\theta(y_w|x) - w(y_l)\log\pi_\theta(y_l|x) \\
% & \quad + w(y_w) - w(y_l) ] \Big) + \eta \beta w(y_w)\log\pi_\theta(y_w|x) \Big)
% \end{aligned}$ \\
% & $\text{where } l^t \text{ is the dynamic label and } w(y) \text{ is the importance weight.}$ \\

\midrule
\textbf{\textit{RQ6: Alignment Tax.}}  \\
\midrule
SPO~\cite{lou2024spomultidimensionalpreferencesequential} & $-\log \sigma \left(\xi_2 \phi_2(x, y_w, y_l) - \xi_1 \phi_1(x, y_w, y_l)\right)$ \\
& $\text{where} \,\,\forall i \in \{1, 2\}\text{,}\,\,\phi_i(x, y_w, y_l) = \log \frac{\pi_i(y_l|x)}{\pi_{i-1}(y_w|x)} - \log \frac{\pi_i(y_w|x)}{\pi_{i-1}(y_l|x)}$ \\
\bottomrule
\end{tabular}
}

\caption{Part variants of DPO objectives discussed in RQs. Method-specific symbols are defined in their respective rows or original papers. Detailed method-specific notation definitions are provided in the Appendix.}
\label{tab:Summary on XPOs}
\end{table*}

In this section, we discuss key research questions on DPO to provide a thorough understanding of its current landscape. We start by comparing DPO with RLHF to highlight the advantages and limitations of DPO. We then investigate the effects of implicit versus explicit reward modeling, focusing particularly on generalization challenges. In addition, we discuss the effects of different feedback and analyze the roles of the KL penalty coefficient and the reference model. Lastly, we review advancements in Online DPO and discuss issues like reward hacking and alignment tax. The part variants of DPO discussed in the RQs are shown in Table ~\ref{tab:Summary on XPOs}.

\begin{figure*}[!hbtp]
    \centering
    \includegraphics[width=0.98\textwidth]{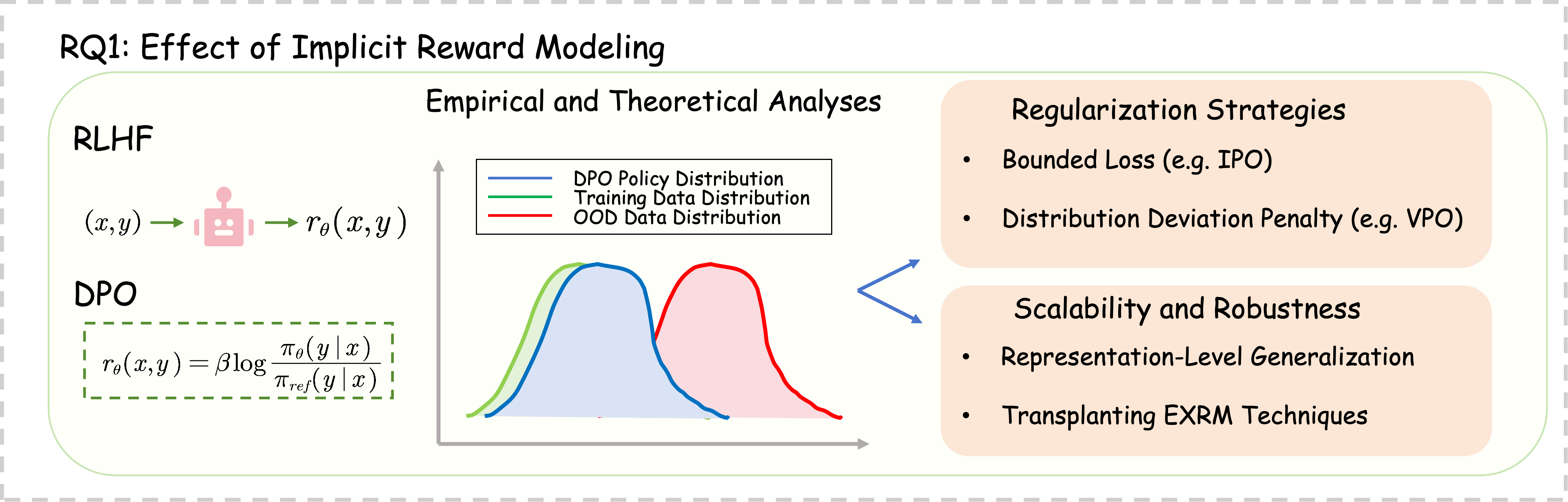}
    \caption{RQ1 Effect of Implicit Reward Modeling. While the DPO policy fits well in the in-distribution training data, its implicit reward mechanism leads to poor generalization on OOD data. To understand and address this key limitation under distribution shifts, recent studies explore implicit reward modeling from three main perspectives: empirical and theoretical analyses, regularization strategies, and scalability and robustness.\del{RQ1 Effect of Implicit Reward Modeling. This figure shows that while the DPO policy distribution fits the in-distribution training data well, its implicit reward mechanism leads to poor generalization. Consequently, its performance degrades significantly on OOD data, revealing a key limitation under distribution shifts compared to explicit reward models.}}
    \label{fig:RQ1}
\end{figure*}

\textbf{RQ0: Why DPO? A Comparison with Classic RLHF.}
To understand why DPO has emerged as an effective alternative to classic RLHF, it is critical to clarify their core advantages and disadvantages. To provide a view spanning from how the models are trained to how they perform in real-world applications, we compare them across six key dimensions: optimization complexity, training cost, reward modeling, OOD generalization, sampling strategy, and optimization granularity. The comparison is summarized in Table~\ref{tab:rlhf_vs_dpo}.

\textbf{Optimization Complexity and Training Cost.} The most direct difference between the two methods lies in how they update the policy. Classic RLHF relies on a complex pipeline that requires training an explicit reward model and running a reinforcement learning algorithm~\cite{ouyang2022training}. This process requires four models to be loaded in memory simultaneously, making it costly and difficult to tune~\cite{ppo}. In contrast, DPO reformulates this process as a simple reward modeling loss. It updates the policy using only two models, which reduces training costs and makes the optimization process more stable~\cite{rafailov2024direct, ethayarajh2024kto}, as shown in Fig. ~\ref{fig:RQ0}.

\textbf{Reward Modeling and OOD Generalization.} However, DPO's simplification in optimization comes at the cost of generalization~\cite{lin2024limitedgeneralizationcapabilityimplicit}. RLHF trains a separate, explicit reward model~\cite{wang2025implicitrewardbridgeunified}, which can learn general human preferences and apply them to OOD datasets, effectively smoothing out noisy labels in the training data~\cite{li2024policyoptimizationrlhfimpact,Jiachen}. DPO, on the other hand, does not use a separate reward model. Instead, its reward is implicitly defined by the policy itself~\cite{wang2025implicitrewardbridgeunified}. Since the policy overfits to the given preference pairs, DPO is highly sensitive to noisy data and performs poorly when faced with OOD prompts that differ from the training distribution~\cite{lin2024limitedgeneralizationcapabilityimplicit, xu2024dpo}.

\textbf{Sampling Strategy and Optimization Granularity.} DPO's reliance on fixed data also highlights another key difference: how they explore and assign credit during training. RLHF is an online method; it actively generates new responses during training. While it receives an instance-level score from the reward model, it uses a value model to assign token-level credit~\cite{tang2024understandingperformancegaponline}. This token-level optimization makes RLHF naturally suited for multi-step reasoning tasks, as it can pinpoint exactly which step went wrong~\cite{lai2024stepdpostepwisepreferenceoptimization}. In contrast, standard DPO is an offline method trained on static datasets~\cite{wang2024offlinerlhfmethodsneed}. It compares entire responses at the instance level. If a long reasoning chain contains a single mistake, DPO penalizes the entire output, which limits its effectiveness in complex tasks~\cite{zeng2024tokenleveldirectpreferenceoptimization, bose2025hybrid}.

\textbf{Applicability.} Considering the aforementioned differences, DPO is optimal for resource-constrained scenarios and well-defined tasks with clear preference boundaries~\cite{rafailov2024direct,xu2024dpo}. In contrast, classic RLHF remains essential for open-ended tasks, long-horizon tasks, and complex reasoning~\cite{bai2022traininghelpfulharmlessassistant, xu2024dpo}.

\rev{\textbf{RQ1: Effect of Implicit Reward Modeling.} 
DPO eliminates the need for training an explicit reward model (EXRM) by reparameterizing the reward function in terms of the optimal policy, as formulated in Eq.~\eqref{ref.4}. This implicit reward mechanism (DPORM), while computationally efficient, faces a critical challenge: it struggles to generalize effectively, especially under distribution shifts~\cite{wang2025implicitrewardbridgeunified,lin2024limitedgeneralizationcapabilityimplicit}. To understand this limitation, recent studies have investigated the implicit reward modeling primarily from three perspectives: \textbf{empirical and theoretical analyses}, \textbf{regularization strategies}, and \textbf{scalability and robustness challenges}~\cite{yan2025dproperties,cen2025valueincentivized,li2024policyoptimizationrlhfimpact}, as illustrated in Fig.~\ref{fig:RQ1}.}

\rev{\textbf{Empirical and Theoretical Analyses.} Regarding the generalization gap, both empirical observations and theoretical analyses highlight the limitations of implicit reward modeling. Lin~\etal~\mbox{\cite{lin2024limitedgeneralizationcapabilityimplicit}} demonstrated that while DPORM achieves comparable performance on in-distribution training data, its performance degrades significantly in OOD scenarios compared to EXRM, with an average accuracy drop of 3\% across five OOD settings. From a theoretical perspective, Yan~\etal~\mbox{\cite{yan2025dproperties}} identified the "3D-Properties" of DPO's implicit reward, proving that a severe gradient imbalance exists between chosen and rejected responses, which restricts its robustness. Furthermore, Li~\etal~\mbox{\cite{li2024policyoptimizationrlhfimpact}} provided error analyses that optimizing policies strictly on offline, in-distribution preference data directly exacerbates these generalization errors.}

\rev{\textbf{Regularization Strategies.} To address the drawbacks of the implicit reward, several methods introduce regularizations directly into the optimization objective. Azar~\etal~\mbox{\cite{azar2023general}} identified the overfitting problem in standard DPO and introduced Identity Preference Optimization (IPO). By replacing the conventional logistic objective with a bounded mean squared error loss, IPO effectively controls the magnitude of the implicit reward, thereby preventing the policy from over-optimizing to the training distribution. Taking a different approach, Cen~\etal~\mbox{\cite{cen2025valueincentivized}} proposed Value-Incentivized Preference Optimization (VPO), which penalizes deviations from a trusted data distribution to suppress offline overfitting and drives exploration in online settings. Beyond data distribution, Wang~\etal~\mbox{\cite{wang2025implicitrewardbridgeunified}} explored improvements to divergence constraints and derived alternative optimization objective using Pearson $\chi^{2}$ and Squared Hellinger divergences; these formulations theoretically preserve the KL constraint during gradient updates, providing a more stable optimization boundary for the implicit reward modeling.}

\rev{\textbf{Scalability and Robustness Challenges.} Despite these advancements, enhancing implicit reward modeling, especially in scalability and robustness, remains an open challenge~\cite{wang2025implicit,lin2024limitedgeneralizationcapabilityimplicit}. Currently, EXRM in standard RLHF benefits from sophisticated generalization strategies at the representation-level. For instance, Yang~\etal~\mbox{\cite{yang2024regularizinghiddenstatesenables}} proposed regularizing hidden states to preserve foundational text-generation capabilities against distribution shifts. While these studies primarily focus on explicit reward modeling, analogous strategies for DPORM remain underexplored. Future research could explore transplanting EXRM-style generalization techniques into the implicit reward formulation to bridge this gap.}

\begin{figure*}[h]
    \centering
    \includegraphics[width=0.98\textwidth]{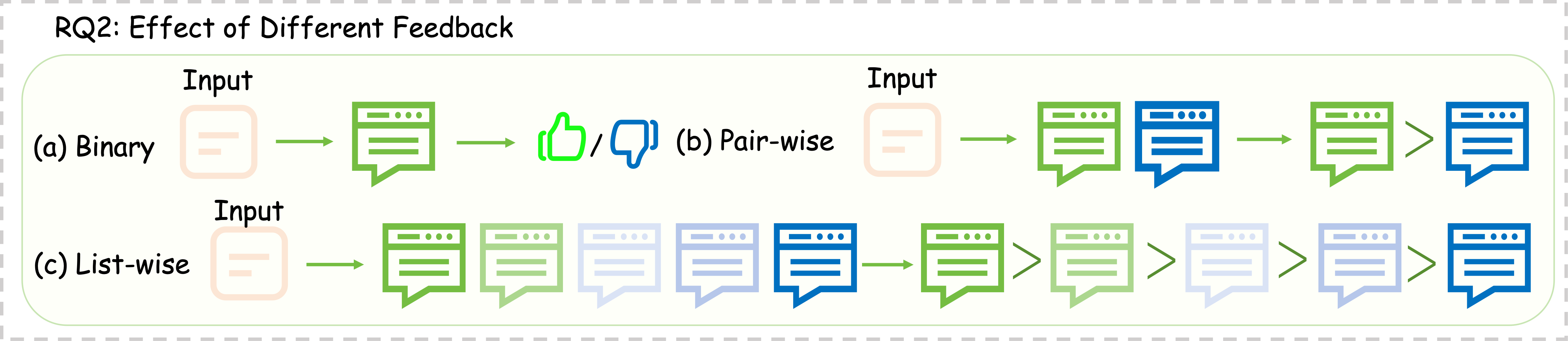}
    \caption{RQ2 Effect of Different Feedback. The figure contrasts three parallel feedback granularities: Binary (absolute judgment), Pair-wise (relative comparison), and List-wise (multi-item ranking).}
    \label{fig:RQ2}
\end{figure*}

\begin{figure}[t]
    \centering
    \includegraphics[width=0.49\textwidth]{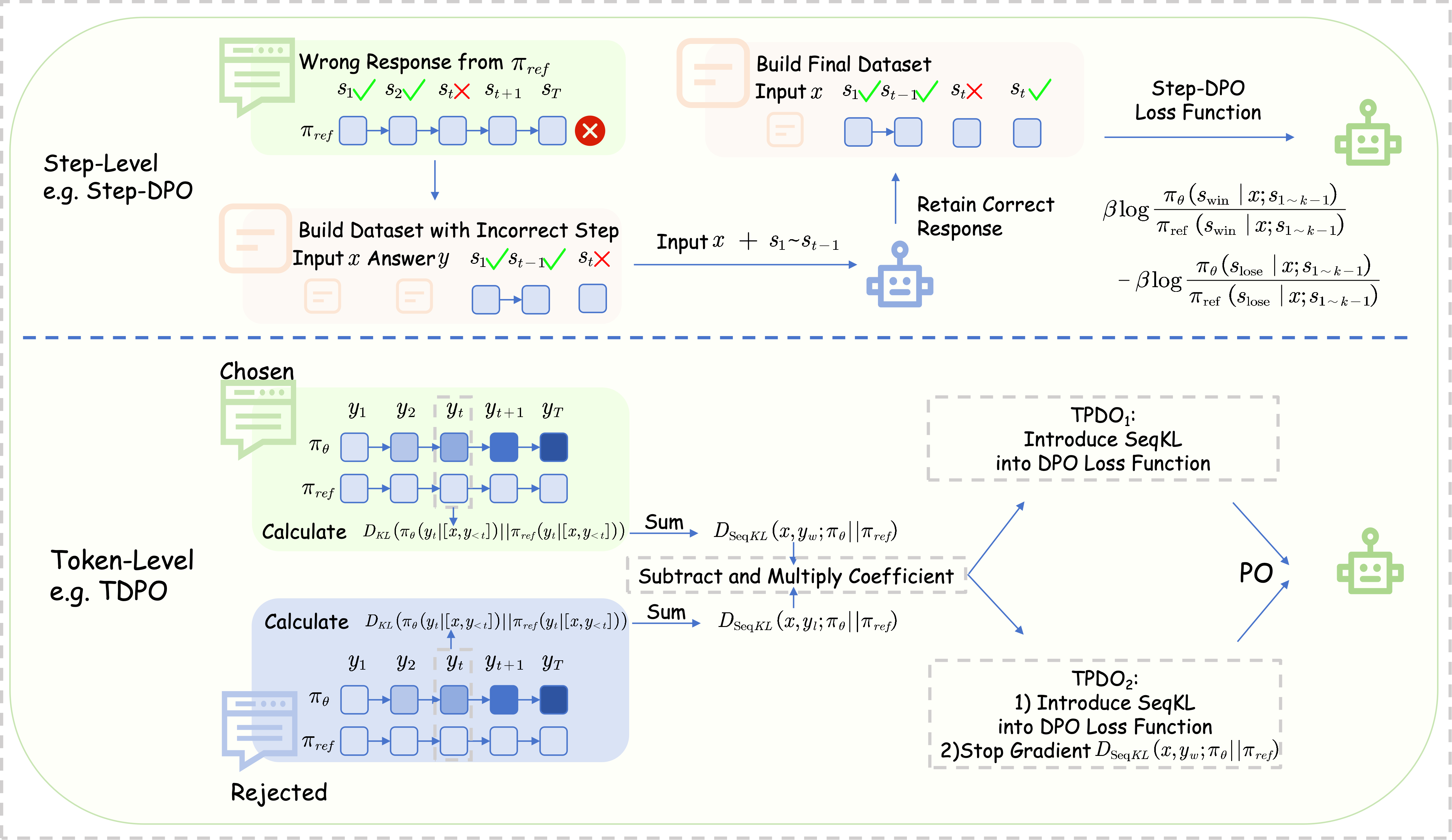}
    \caption{Step-Level and Token-Level Feedback Methods. The upper part shows Step-DPO, which provides step-level supervision by focusing the optimization unit on the first erroneous reasoning step. The lower part depicts TDPO, which performs finer-grained policy optimization by applying a per-token KL divergence constraint.}
    \label{fig:RQ2_step&token}
\end{figure}

\rev{\textbf{RQ2: Effect of Different Feedback.}}
\rev{Standard DPO relies on a highly constrained form of feedback: it is pairwise in format, instance-level in granularity, and ordinal rather than cardinal in its feedback signal. However, pairwise annotation is costly and difficult to scale~\cite{ethayarajh2024kto}; instance-level optimization provides no credit assignment for localized errors within a response~\cite{zeng2024tokenleveldirectpreferenceoptimization}; and relying solely on ordinal feedback provides no absolute quality signal~\cite{zeng2024tokenleveldirectpreferenceoptimization, azar2023general}. To fully exploit the potential of preference data, recent studies have expanded the concept of feedback across three distinct dimensions: exploring alternative feedback formats, refining feedback granularity, and introducing dynamic or hybrid feedback signals, as demonstrated in Fig.~\ref{fig:RQ2} and Fig.~\ref{fig:RQ2_step&token}.}

\rev{\textbf{Alternative Feedback Formats.} Moving beyond the strict pair-wise limitation, researchers have explored diverse feedback structures to improve data efficiency. At the binary level, KTO~\mbox{\cite{ethayarajh2024kto}} eliminates the need for paired preferences. Inspired by prospect theory~\cite{Tversky1992AdvancesIP}, it operates on binary feedback, simply judging whether a single response is desirable or undesirable. Specifically, rather than comparing a pair, KTO measures each response against a KL-based baseline and applies asymmetric weights to desirable and undesirable outcomes, reflecting the human tendency to overweight losses relative to gains. Beyond binary feedback, scaling up to multi-candidate feedback, RAFT~\mbox{\cite{dong2023raft}} and RRHF~\mbox{\cite{yuan2023rrhfrankresponsesalign}} exploit list-wise ranking feedback, training the policy to assign probabilities proportional to the explicit ranking scores of multiple responses. For more complex reasoning environments, ULTRAINTERACT~\mbox{\cite{yuan2024advancing}} constructs tree-structured feedback, organizing multi-turn correct and incorrect actions into logical branches. Furthermore, M-DPO~\mbox{\cite{xiong2025building}} adapts feedback for tool-use environments by masking out external observations, ensuring the reward signal focuses strictly on the agent's internal reasoning steps rather than uncontrollable environmental messages.}

\rev{\textbf{Finer Feedback Granularities.} Standard DPO applies feedback at the instance level, treating an entire response as an action. However, this holistic feedback lacks the fine-grained credit assignment, e.g., identifying localized errors in long reasoning chains. Recognizing that DPO can be theoretically mapped to a token-level MDP~\mbox{\cite{rafailov2024rqlanguagemodel}}, subsequent works have sought to provide denser supervision. Step-DPO~\mbox{\cite{lai2024stepdpostepwisepreferenceoptimization}} shifts the optimization unit from the whole response to an individual reasoning step, providing step-level feedback that precisely targets the first erroneous action. Furthermore, in token-level feedback, TDPO~\mbox{\cite{zeng2024tokenleveldirectpreferenceoptimization}} applies per-token KL divergence constraints to improve localized alignment. Acknowledging that token importance fluctuates, TI-DPO~\mbox{\cite{yang2026tokenimportance}} introduces a hybrid weighting mechanism, namely gradient attribution and a Gaussian prior, to quantify token significance, while cDPO~\mbox{\cite{lin2024critical}} delivers targeted feedback by applying dynamic penalty weights exclusively to critical negative tokens.}

\rev{\textbf{Dynamic and Hybrid Feedback Signals.} The standard DPO feedback solely penalizes the relative reward difference between two responses, which empirically causes the absolute likelihood of chosen responses to decrease over training~\mbox{\cite{pal2024smaug, rafailov2024rqlanguagemodel}}. To correct this, recent works have transformed the feedback mechanism from purely relative to a hybrid signal. Several works~\mbox{\cite{pang2024iterativereasoningpreferenceoptimization,yuan2023rrhfrankresponsesalign,xu2024contrastive}} blend the relative preference reward with absolute feedback by adding a negative log-likelihood (NLL) or behavior cloning (BC) regularizer, forcing the model to increase the probability of preferred data. Furthermore, beyond static, pre-collected human feedback, recent methodologies have pioneered dynamic feedback generation via self-play and Nash learning~\mbox{\cite{wu2024selfplaypreferenceoptimizationlanguage,rosset2024directnashoptimizationteaching,munos2024nashlearninghumanfeedback}}. For example, SPIN~\mbox{\cite{chen2024self}} formulates alignment as a two-player game where the model derives iterative, self-generated feedback by continuously competing against its own past generations, effectively eliminating external static reward models and preference data.}

\begin{figure*}[h]
    \centering
    \includegraphics[width=0.98\textwidth]{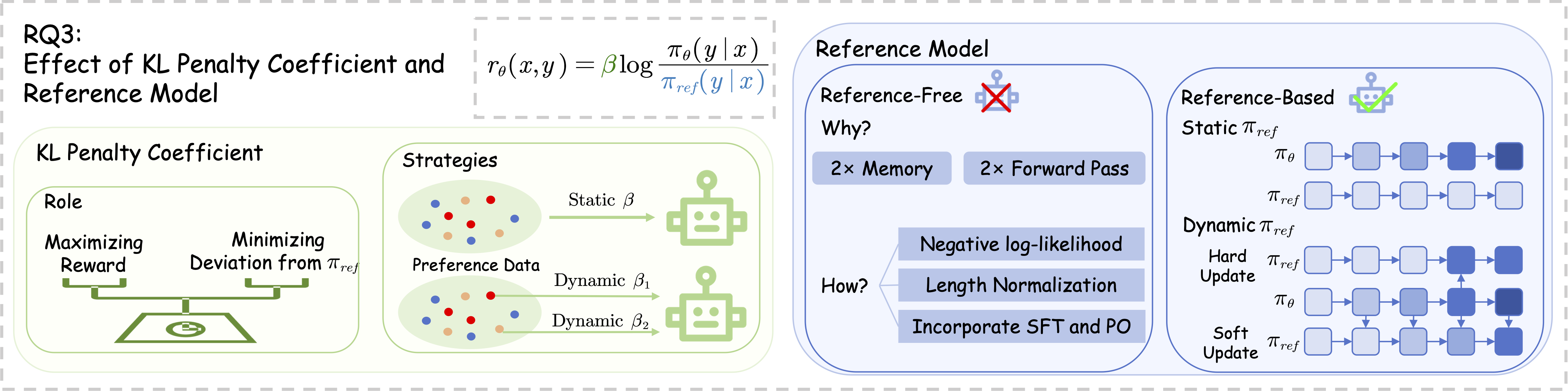}
    \caption{RQ3 Effect of KL Penalty Coefficient and Reference Model. The left panel illustrates the key role of the KL penalty in controlling the trade-off between reward maximization and policy deviation, showcasing both static (fixed coefficient) and dynamic (adaptive coefficient) strategies. The right panel categorizes methods based on their reliance on a reference model into two main classes: reference-based and reference-free approaches.}
    \label{fig:RQ3}
\end{figure*}

\textbf{RQ3: Effect of KL Penalty Coefficient and Reference Model.} As discussed in RQ1, DPO implicitly parameterizes the reward model using a reference policy $\pi_{\mathrm{ref}}$ and a KL penalty coefficient $\beta$. This implicit reward modeling in Eq.~\eqref{ref.4} establishes the critical trade-off between maximizing preference rewards and minimizing policy deviation. However, relying on a static coefficient may lead to over-penalization or under-penalization of policy updates~\mbox{\cite{liu2024understandingreferencepoliciesdirect,wu2024betadpodirectpreferenceoptimization}}, while maintaining a fixed reference model also incurs significant additional memory overhead and risks the policy exploiting OOD regions~\mbox{\cite{xu2024contrastive,xu2024dpo}}. To address these limitations, recent studies have advanced DPO across two primary directions: dynamically calibrating the KL penalty~\cite{liu2024understandingreferencepoliciesdirect,wu2025alphadpo}, and exploring strategies to eliminate or dynamically update the reference model~\cite{feng2024analyzingunderstandinglimitationsdpo,gorbatovski2025learn}, as depicted in Fig. ~\ref{fig:RQ3}.

\rev{\textbf{Dynamic Calibration of the KL Penalty.} In standard DPO, a static $\beta$ controls the regularization strength. Liu~\etal~\mbox{\cite{liu2024understandingreferencepoliciesdirect}} explored this static constraint, revealing that while a smaller KL penalty generally improves performance, excessively low values cause the model to assign drastically different probabilities to a small subset of specific tokens, leading to catastrophic degradation. To overcome the drawbacks of static tuning, several influential works have proposed dynamic calibration mechanisms. For instance, Wu~\etal~\mbox{\cite{wu2024betadpodirectpreferenceoptimization}} introduced $\beta$-DPO, which dynamically calibrates $\beta$ at the batch level. Rather than treating all preference pairs equally, $\beta$-DPO mathematically adjusts the penalty based on the implicit reward margin between the chosen and rejected responses, explicitly filtering outliers and preventing overfitting to noisy data. Extending this adaptive philosophy, Lee~\etal~\mbox{\cite{lee2025kl}} proposed $\epsilon$-DPO to adjust instance-level KL penalties by observing the monotonicity of log-likelihood ratios under perturbed $\beta$ values, while Wu~\etal~\mbox{\cite{wu2025alphadpo}} developed $\alpha$-DPO to calibrate the reward margin itself using an adaptive preference distribution.}

\rev{From a more theoretical perspective, Xie~\etal~\mbox{\cite{xie2025exploratory}} provided an insight that DPO implicitly performs a form of $Q^*$-function approximation. Building on this observation, they proposed Exploratory Preference Optimization (XPO) specifically for online settings. XPO augments the standard DPO objective with an exploration bonus derived from the "optimism in the face of uncertainty" principle in reinforcement learning. This explicitly encourages the policy to explore novel behaviors beyond the initial reference policy's support, effectively addressing the sample inefficiency and "passive exploration" limitations in static-penalty DPO.}

\rev{\textbf{Eliminating vs. Dynamic Updating the Reference Model.} A standard DPO pipeline requires double the memory and forward passes due to the reference model, and it heavily relies on a separate SFT warm-up stage~\mbox{\cite{feng2024analyzingunderstandinglimitationsdpo}}. Consequently, a line of research advocates for its \textit{elimination}. Xu~\etal~\mbox{\cite{xu2024contrastive}} proposed Contrastive Preference Optimization (CPO), which mathematically derives an upper bound of the DPO loss to drop the reference model entirely. To prevent the policy from deviating too far from the target distribution without the reference anchor, CPO explicitly introduces a behavior cloning regularizer. Similarly, addressing the training burden, Hong~\etal~\mbox{\cite{hong2024orpomonolithicpreferenceoptimization}} introduced ORPO, a monolithic framework that bypasses both the SFT warm-up and the reference model. ORPO achieves this by integrating an odds ratio-based penalty directly into the NLL, dynamically penalizing rejected responses. Following a different intuition, Meng~\etal~\mbox{\cite{meng2024simpo}} proposed SimPO, which directly utilizes the average log-probability of a sequence as the implicit reward, eliminating the reference model while simultaneously introducing a length-normalized margin to counteract verbosity.}

\rev{In contrast, another line argues for the \textit{preservation and dynamic updating} of the reference model to ensure training stability. Liu~\etal~\mbox{\cite{liu2024understandingreferencepoliciesdirect}} demonstrated that the KL constraint anchoring the policy to the reference model is a necessary regularization: without it, the policy is prone to severe model degeneration caused by extreme probability assignments. However, static reference models remain vulnerable to OOD exploitation~\mbox{\cite{xu2024dpo}}. To balance training stability with policy flexibility, Gorbatovski~\etal~\mbox{\cite{gorbatovski2025learn}} proposed TR-DPO. Instead of removing the reference model, TR-DPO introduces a trust-region approach that dynamically updates the reference policy during training via soft (parameter interpolation) and hard (periodic replacement) updates. This allows for safe and continuous policy deviation while avoiding over-optimization.}

\begin{figure*}[t]
    \centering
    \includegraphics[width=0.98\textwidth]{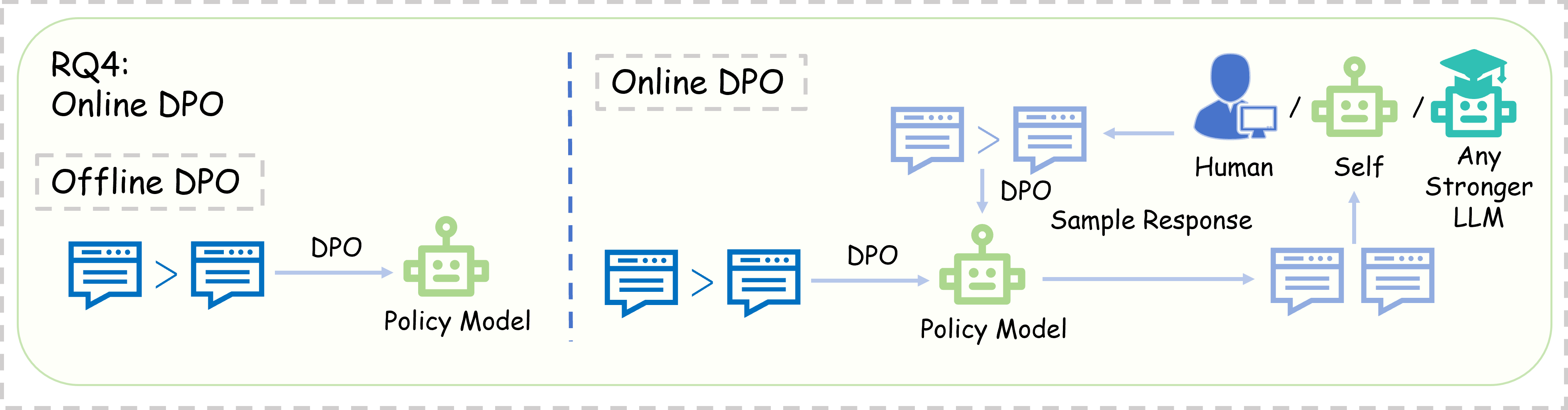}
    \caption{RQ4 Online DPO. This figure shows the training processes for Offline DPO and Online DPO. In the Online DPO, newly sample responses from each iteration can be annotated through various means, including by human annotators, the model itself, or any other stronger model.}
    \label{fig:RQ4}
\end{figure*}

\begin{figure}[h]
    \centering
    \includegraphics[width=0.49\textwidth]{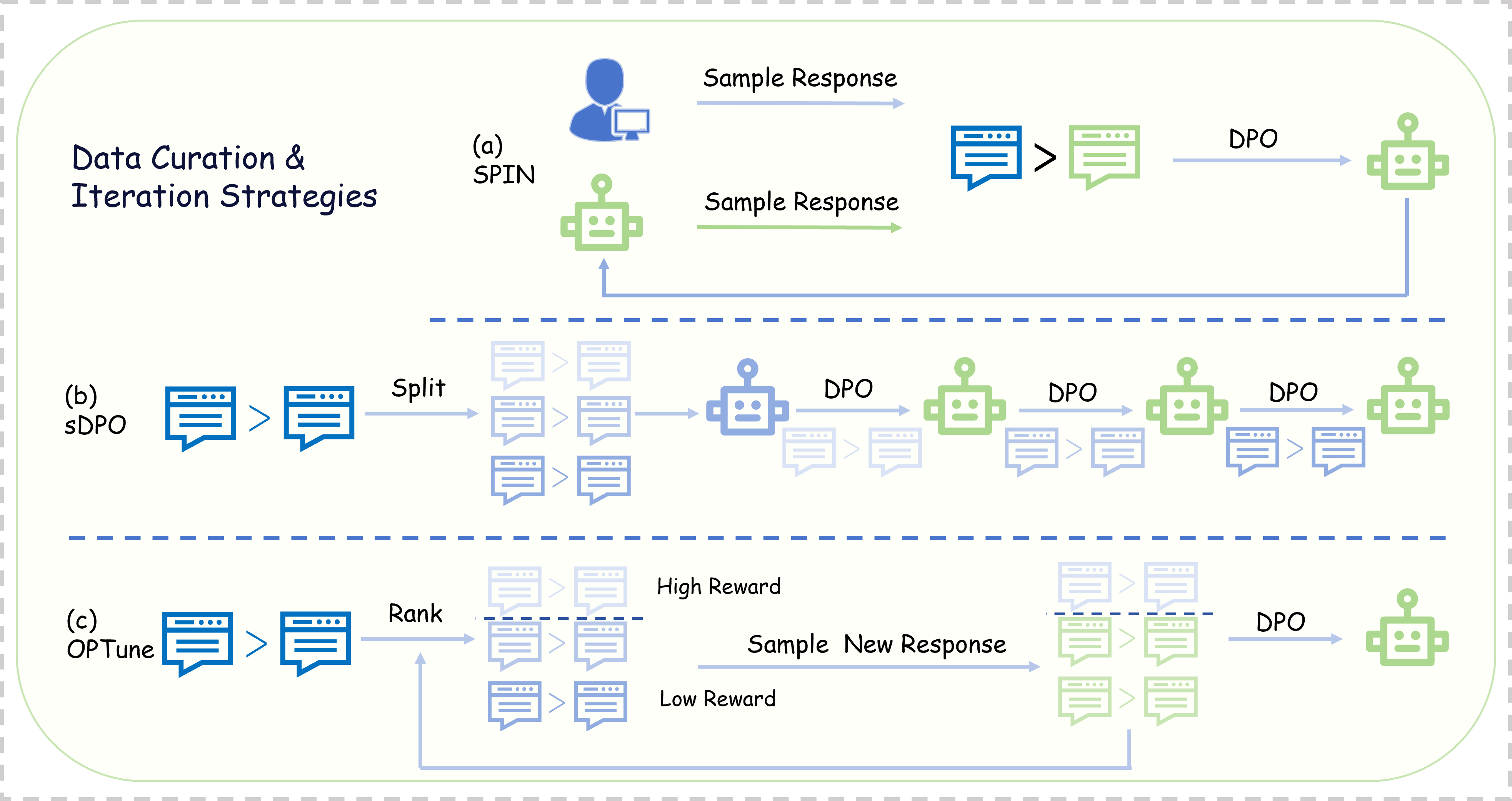}
    \caption{Data Curation and Iteration Strategies in Online DPO. The panel (a) shows a self-play framework that iteratively optimizes by pairing model responses against human responses. The panel (b) shows sDPO, which trains on data partitions with a dynamically updated reference model. The panel (c) shows the OPTune algorithm, which combines selective data regeneration with a weighted-DPO loss to accelerate optimization.}
    \label{fig:RQ4_data}
\end{figure}

\textbf{RQ4: Online DPO.} Standard DPO is typically an offline alignment framework~\cite{rafailov2024direct}, and it suffers from distributional discrepancy between the training data and the current policy~\cite{ren2025learning}, resulting in a performance gap compared to online reinforcement learning methods~\cite{tang2024understandingperformancegaponline,wang2024offlinerlhfmethodsneed}. To explicitly address this limitation, recent research transitions DPO into online and iterative paradigms, addressing two core bottlenecks: efficiently constructing online preference data to mitigate distributional discrepancy~\cite{guo2024controllablepreferenceoptimizationcontrollable,chen2024self}, and designing robust iterative optimization strategies to improve training dynamics~\cite{qi2024onlinedpoonlinedirect,bose2025hybrid}, as shown in Fig.~\ref{fig:RQ4} and Fig.~\ref{fig:RQ4_data}.

\textbf{Efficient Online Data Curation.} Online DPO requires continuous data collection, making costly human annotation a critical bottleneck. To resolve this, recent works heavily leverage AI feedback mechanisms, from external model reliance to self-rewarding. For example, OAIF~\cite{guo2024directlanguagemodelalignment} mitigates distribution shifts by using external, stronger LLMs as surrogate human judges. To break the dependency on proprietary external models, Yuan~\etal~\cite{yuan2024self} proposed the Self-Rewarding framework, where the policy internalizes the evaluation process; following an "Instruction Following" stage, the model acts as its own judge across five metrics (relevance, coverage, usefulness, clarity, and expertise) to construct online preference pairs. Taking a different approach, PCO~\cite{xu2024thingscringeothersiterative} relies on an explicit reward model to label generated responses, generalizing the Cringe Loss~\cite{adolphs2022cringelosslearninglanguage} from binary to pairwise feedback using a soft-margin extension. Pushing toward self-sufficiency by bypassing explicit reward scoring, SPIN~\cite{chen2024self} introduces an iterative self-play framework that constructs preference pairs by treating the model's online generations as rejected responses and human ground-truths as chosen responses, achieving alignment through the model competing against its own past iterations.

Beyond data annotation, online data generation introduces severe computational bottlenecks. Consequently, recent studies optimize data curation by shifting the focus from how to annotate to what to sample~\cite{chen2024optuneefficientonlinepreference,shi2025the}. For instance, observing that generation accounts for approximately 70\% of training time, OPTune~\cite{chen2024optuneefficientonlinepreference} selectively regenerates only low-reward responses and utilizes a weighted DPO (w-DPO) objective to assign greater weight to pairs with larger reward gaps. Further advancing sample efficiency, ActiveDPO~\cite{lin2026activedpo} explicitly accounts for the influence of the LLM on data selection, actively sampling the most informative preference pairs based on non-linear reward functions rather than randomly exploring the state space.

\textbf{Iterative Optimization Strategies.} With online data generation, directly optimizing the policy against a static reference model using standard DPO causes the KL divergence to be unstable~\cite{kim2024sdpo,gorbatovski2025learn}. To establish a basic baseline for reference model updating, Kim~\etal~\cite{kim2024sdpo} proposed sDPO, which partitions offline datasets and iteratively places the policy model of the current iteration as the reference model of the next iteration to set tighter optimization bounds. Beyond reference updating, to maintain gradient update momentum during continuous training, OFS-DPO~\cite{qi2024onlinedpoonlinedirect} introduces an alternating fast-slow dual-LoRA architecture, which swaps the roles of the fast and slow modules at specific steps and applies a novel regularization term. To explicitly encourage policy exploration, HPO~\cite{bose2025hybrid} theoretically establishes sample complexity bounds by relaxing concentrability conditions on offline data, integrating an optimistic regularizer to balance offline data and online exploration via a tuning parameter $\gamma$. 

Furthermore, prolonged DPO training risks text degeneration~\cite{ren2025learning}, even iterative optimization can degrade specific capabilities like reasoning~\cite{pang2024iterativereasoningpreferenceoptimization}. To mitigate this, IRPO~\cite{pang2024iterativereasoningpreferenceoptimization} applies a NLL to chosen responses, demonstrating that generating new pairs and retraining from the previously trained model improves reasoning performance steadily until saturation. Advancing this stability in complex domains, TRPA~\cite{trpa_dpo} enforces trust-region constraints to guarantee monotonic improvements, achieving reasoning performance competitive with RLHF. Finally, the EVOLVE framework~\cite{evolving_self_refine} proposes a two-stage iterative approach: it first trains the model to follow correction instructions, and then uses the model to iteratively revise its own generated text. These revisions are collected as preference data for subsequent training to improve the model's self-correction ability.

\begin{figure*}[t]
    \centering
    \includegraphics[width=0.98\textwidth]{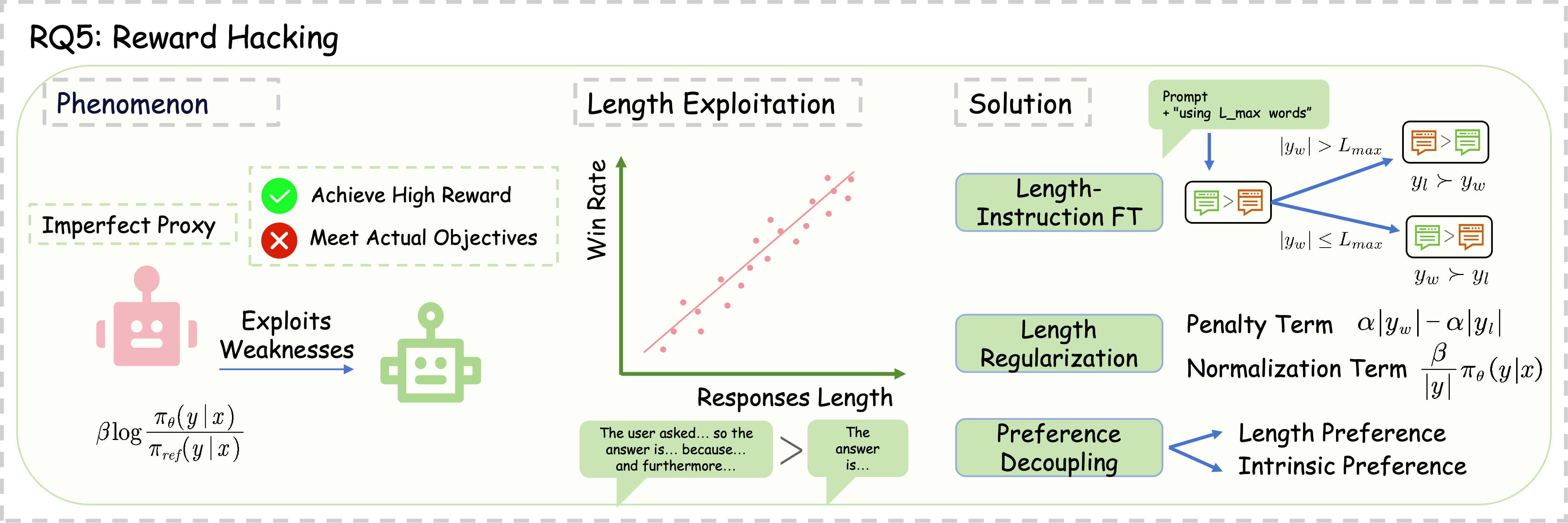}
    \caption{RQ5 Reward Hacking. Reward hacking occurs when a policy exploits the limited generalization of its reward model to achieve a high reward without meeting the actual objectives. A prominent example is length exploitation, where models generate longer responses to exploit data biases, often at the cost of quality. This figure shows various mitigation strategies to counter this issue, including Length-Instruction FT, length regularization, and preference decoupling.}
    \label{fig:RQ5}
\end{figure*}

\textbf{RQ5: Reward Hacking.}
Reward hacking emerges when a policy model achieves high proxy rewards without fulfilling actual alignment objectives~\cite{rashidinejad2025sail,park-etal-2024-disentanglinglength}. As discussed in RQ1, DPO's implicit reward modeling tends to overfit the in-distribution training data, resulting in poor OOD generalization. Consequently, the policy is more susceptible to obtaining high rewards through OOD exploitation. \del{In DPO, the implicit reward mechanism makes the policy more susceptible to obtaining artificially high rewards through OOD exploitation.} A prominent example is \textit{length exploitation}~\cite{dubois2024alpacafarmsimulationframeworkmethods, singhal2023longwaygoinvestigating, Kabir_2024}, wherein models generate increasingly verbose responses that may contain hallucinated content to obtain higher rewards~\cite{wang2023farcamelsgoexploring,xiao2024detectingmitigatinghallucinationlarge}. To address these vulnerabilities, we categorize recent mitigation strategies into two directions: mitigating length exploitation and mitigating other OOD exploitation in DPO, as illustrated in Fig.~\ref{fig:RQ5}.

\rev{\textbf{Mitigating Length Exploitation.} To address the prevalent issue of length bias, most works focus on modifying the optimization objective to penalize or decouple verbosity. Park~\etal~\mbox{\cite{park-etal-2024-disentanglinglength}} linked length exploitation to OOD bootstrapping and introduced R-DPO, which adds a straightforward length-based penalty to the regularization term. Taking a more structural approach, SimPO~\mbox{\cite{meng2024simpo}} incorporates length normalization directly into the reward difference formulation to align the training objective with the generation length normalization. From the theoretical mechanics, Liu~\etal~\mbox{\cite{liu2024lengthdesensitizationdirectedpreference}} uncovered an underlying link between DPO's implicit reward and data length, proposing LD-DPO to mathematically decouple explicit length preference from intrinsic preferences. In contrast, taking a data-centric approach rather than modifying the objective, Yuan~\etal~\mbox{\cite{yuan2024followinglengthconstraintsinstructions}} introduced LIFT-DPO. This data-level intervention incorporates length constraint instructions into general datasets and explicitly rejects chosen responses that exceed specified length limits, improving length instruction-following without degrading general capabilities.}

\rev{\textbf{Mitigating Other OOD Exploitation in DPO.} Broadening the scope beyond length bias, recent works target general reward hacking from two perspectives: data coverage and optimization dynamics~\cite{rashidinejad2025sail,yan2025dproperties}. From the data perspective, Rashidinejad~\etal~\mbox{\cite{rashidinejad2025sail}} identified the over-optimization of OOD actions caused by poor high-reward data coverage. To address this limitation, they proposed POWER-DL, combining weighted entropy regularization to enforce pessimism on uncovered actions with dynamic label weighting to prevent destructive policy updates. Complementing this, Yan~\etal~\mbox{\cite{yan2025dproperties}} examined the optimization dynamics, and revealed that DPO implicitly hacks the reward by excessively minimizing the likelihood of rejected responses, which forces probability mass to redistribute into unpredictable OOD regions. To mitigate this issue, they introduced adaptive gradient weighting to down-weight the gradients for excessively low-likelihood rejected responses and incorporate NLL to regularize the model toward the reference policy.}

\begin{figure*}[t]
    \centering
    \includegraphics[width=0.98\textwidth]{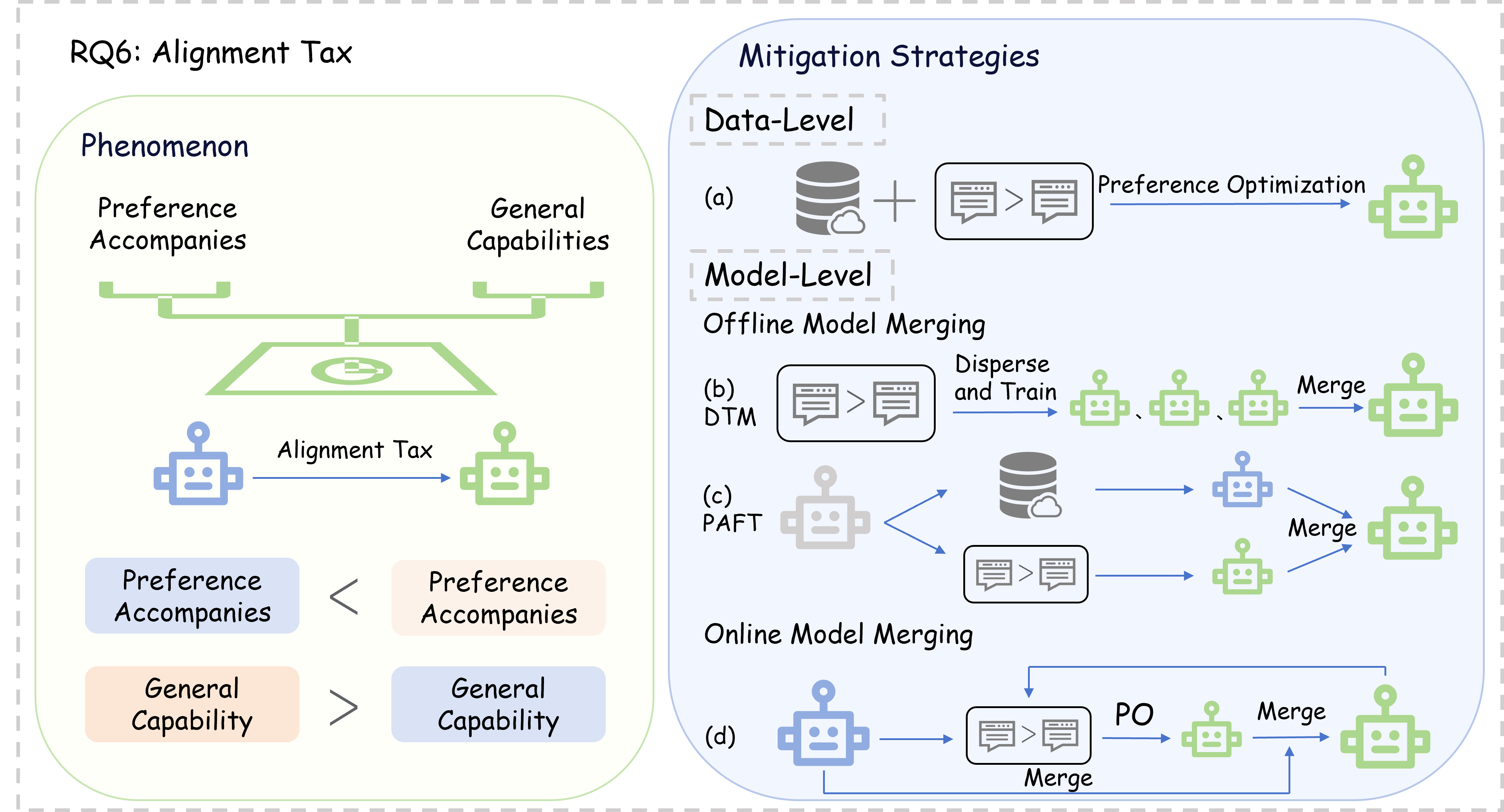}
    \caption{RQ6 Alignment Tax. The scale in the left panel illustrates the "alignment tax"---the trade-off between \rev{preference alignment} and general capabilities. The right panel outlines various strategies to mitigate this issue, primarily focusing on data-level and model-level strategies.}
    \label{fig:RQ6}
\end{figure*}

\textbf{RQ6: Alignment Tax.} 
Both RLHF and DPO aim to align models with human preferences. However, previous studies have identified a phenomenon known as the  \textit{alignment tax}~\cite{ouyang2022training,openai2024gpt4}, where improved \rev{preference alignment} often accompanies degraded performance in other tasks or dimensions. To address this issue, recent studies have proposed various methods to reduce its effect~\cite{tu2023sighttextmultimodaltraining,anthropic2024claude,wu2024continuallearning}, which can be broadly categorized into recovering general capabilities, balancing multiple objectives, and restoring model calibration, as illustrated in Fig.~\ref{fig:RQ6}.

\textbf{Recovering General Capabilities.} 
The most common alignment tax of language models is the performance drop on NLP tasks~\cite{ouyang2022training}. Kim~\etal~\cite{kim2024rethinkingroleproxyrewards} observed that the key neurons for safety and helpfulness significantly overlap but require different activation patterns, which may explain this capability trade-off. To mitigate this, Ouyang~\etal~\cite{ouyang2022training} proposed incorporating pretraining data into the RLHF fine-tuning process. Furthermore, Fu~\etal~\cite{fu2024dispersethenmergepushinglimitsinstruction} found that the alignment tax is also present during the SFT stage due to data biases. They proposed to disperse the training data to train multiple sub-models, and finally merge them to reduce the alignment tax. 
Similarly, model merging has been widely explored to balance base and aligned models. Lin~\etal~\cite{lin2024mitigatingalignmenttaxrlhf} explored offline model averaging, which interpolates between pre- and post-RLHF model weights to achieve a more efficient reward-tax Pareto front. For online optimization, Lu~\etal~\cite{lu2024onlinemergingoptimizers} online merged the gradients of the policy model with the SFT model at each optimization step. Pentyala~\etal~\cite{pentyala2024paftparalleltrainingparadigm} introduced PAFT, which independently performs SFT and preference alignment on respective datasets before fusing the parameters.

\textbf{Balancing Multiple Objectives.} 
Another perspective on the alignment tax is the trade-off among different alignment objectives (e.g., helpfulness vs. safety). Lou~\etal~\cite{lou2024spomultidimensionalpreferencesequential} found that the alignment tax will accumulate in the iterative fine-tuning of each dimension, potentially leading to catastrophic model collapse. They proposed Sequential Preference Optimization (SPO) with additional constraints to prevent model performance degradation. To explicitly balance multiple objectives, Guo~\etal~\cite{guo2024controllablepreferenceoptimizationcontrollable} introduced Controllable Preference Optimization, which assigns explicit preference scores to different objectives to achieve Pareto improvements. Similarly, Zhou~\etal~\cite{zhou2024beyond} proposed MODPO, which includes a dynamic margin in the DPO loss to steer language models across diverse preferences.

\textbf{Restoring Model Calibration.} 
Beyond aforementioned trade-offs, DPO also introduces an alignment tax on model calibration, causing models to be overconfident and degrading their reliability~\cite{xiao2025restoring,kim2025dpopro}. To address this issue, they proposed RCFT~\cite{xiao2025restoring}, which integrates dynamic confidence adjustment during fine-tuning and temperature scaling during inference, alleviating this specific alignment tax. On the data optimization side, Kim~\etal~\cite{kim2025dpopro} proved that since DPO maximizes the margins in weak preference data, it forces the model to be confident in noisy signals. To this end, they proposed an optimization regularizer in the DPO objective, penalizing confident updates on weak preference.

\section{Datasets}
\label{ch: Datasets}

\begin{table*}[!htbp]
\centering
\resizebox{0.95\textwidth}{!}{%
\begin{tabular}{llcccc}
\toprule
\textbf{Datasets} & \textbf{Task Description} & \textbf{Modality} & \textbf{Venue} & \textbf{Feedback Granularities} & \textbf{Dataset Size (\#Rows)} \\ 
\midrule
\multicolumn{6}{c}{\textbf{Human Labeled}} \\
\midrule
% --- Text ---
Alpaca Farm Human~\cite{dubois2024alpacafarmsimulationframeworkmethods} & Instruction Tuning & Text & NeurIPS 2023 & Pair-wise & 101,000 \\
HH-RLHF~\cite{bai2022traininghelpfulharmlessassistant} & Preference Tuning & Text & - & Pair-wise & 169,352 \\
Nvidia/HelpSteer~\cite{wang2023helpsteermultiattributehelpfulnessdataset} & Preference Tuning & Text & NAACL 2024 & Point-wise & 37,120 \\
Nvidia/HelpSteer2~\cite{wang2024helpsteer2opensourcedatasettraining} & Preference Tuning & Text & NeurIPS 2024 & Point-wise & 21,362 \\
Nvidia/HelpSteer3~\cite{wang2025helpsteer3preferenceopenhumanannotatedpreference} & Preference Tuning & Text & ACL 2025 & Pair-wise & 40,476 \\
OpenAI:Webgpt~\cite{nakano2022webgptbrowserassistedquestionansweringhuman} & Question Answering & Text & - & Pair-wise & 19,578 \\
OpenAssistant/oasst1~\cite{openassistant} & Conversation & Text & NeurIPS 2023 & List-wise & 88,838 \\
Stanfordnlp/SHP~\cite{pmlr-v162-ethayarajh22a} & Preference Tuning & Text & ICML 2022 & Pair-wise & 385,563 \\
Summarize from Human Feedback~\cite{stiennon2022learning} & Summarization & Text & NeurIPS 2020 & Pair-wise & 193,841 \\
Tasksource/Dpo-pairs~\cite{sileo-2024-tasksource-large} & Preference Tuning & Text & LREC-COLING 2024 & Pair-wise & 5,128,939 \\
% --- Multimodal ---
MM-RLHF~\cite{zhang2025mmrlhfstepforwardmultimodal} & Preference Tuning & Multimodal & ICML 2025 & List-wise & 16,342 \\
MMSafe-PO~\cite{li2025harmlessmultimodalassistantsblind} & Safety Preference & Multimodal & ACM MM 2025 & Pair-wise & 5,667 \\
RLHF-V-Dataset~\cite{rlhf-v} & Preference Tuning & Multimodal & CVPR 2024 & Pair-wise & 5,733 \\

\midrule
\multicolumn{6}{c}{\textbf{AI Labeled}} \\
\midrule
% --- Text ---
Alpaca Farm GPT-4~\cite{dubois2024alpacafarmsimulationframeworkmethods} & Instruction Tuning & Text & NeurIPS 2023 & Pair-wise & 20,000 \\
ANAH~\cite{gu2025maskdpogeneralizablefinegrainedfactuality} & Hallucination Detection & Text & ICLR 2025 & Point-wise & 783 \\
Chatml-dpo-pairs~\cite{mukherjee2023orca} & Instruction Tuning & Text & - & Pair-wise & 12,859 \\
CodeSteer-DPO-Dataset~\cite{chen2025codesteersymbolicaugmentedlanguagemodels} & Code Generation & Text & ICML 2025 & Pair-wise & 4,462 \\
Distilabel-capybara-dpo~\cite{ref:capybara} & Conversation & Text & - & Pair-wise & 7,563 \\
Gutenberg-DPO~\cite{ref:jondurbin/gutenberg-dpo-v0.1} & Creative Writing & Text & - & Pair-wise & 918 \\
Imdb-dpo~\cite{yuasosnin/imdb-dpo} & Summarization & Text & - & Pair-wise & 10,000 \\
Magpie-Llama-3.1-DPO~\cite{xu2024magpiealignmentdatasynthesis} & Instruction Tuning & Text & ICLR 2025 & Pair-wise & 100,000 \\
Math-Step-DPO~\cite{lai2024stepdpostepwisepreferenceoptimization} & Mathematical Reasoning & Text & - & Pair-wise & 10,795 \\
Nectar~\cite{starling2023} & Preference Tuning & Text & COLM 2024 & List-wise & 182,954 \\
Py-dpo~\cite{ref:jondurbin/py-dpo-v0.1} & Code Generation & Text & - & Pair-wise & 9,466 \\
Stack Exchange Preference~\cite{h4stackexchange} & Instruction Tuning & Text & - & Point-wise & 10,741,532 \\
Tldr-preference-dpo~\cite{ref:dataset-tldr-preference-dpo} & Summarization & Text & - & Pair-wise & 522 \\
Truthy-dpo~\cite{ref:jondurbin/truthy-dpo-v0.1} & Truthfulness Preference & Text & - & Pair-wise & 1,016 \\
UltraFeedback~\cite{ultrafeedback} & Preference Tuning & Text & ICML 2024 & List-wise & 63,967 \\
Zsql-postgres-dpo~\cite{ref:zerolink/zsql-postgres-dpo} & Text-to-SQL & Text & - & Pair-wise & 259,326 \\
% --- Multimodal ---
HSA-DPO~\cite{xiao2024detectingmitigatinghallucinationlarge} & Preference Tuning & Multimodal & AAAI 2025 & Pair-wise & 8,386 \\ 
MM-IFDPO-23k~\cite{ding2025mmifenginemultimodalinstructionfollowing} & Instruction Tuning & Multimodal & ICCV 2025 & Pair-wise & 22,555 \\
MMInstruction/VLFeedback~\cite{silkie} & Instruction Tuning & Multimodal & EMNLP 2024 & List-wise & 80,258 \\
RLAIF-V-Dataset~\cite{yu2024rlaifvaligningmllmsopensource} & Preference Tuning & Multimodal & CVPR 2025 & Pair-wise & 33,835 \\
sqrti/SPA-VL~\cite{zhang2024spavlcomprehensivesafetypreference} & Safety Preference & Multimodal & CVPR 2025 & Pair-wise & 100,788 \\
Unified Preference Dataset~\cite{wang2025unified} & Preference Tuning & Multimodal & \rev{NeurIPS 2025}& Point/Pair-wise & 236,000 \\

\midrule

\multicolumn{6}{c}{\textbf{Evaluation Benchmarks}} \\
\midrule
% --- Text & Multimodal ---

Chatbot Arena~\cite{zheng2023judgingllmasajudgemtbenchchatbot} & Model Evaluation & Text & NeurIPS 2023 & Pair-wise & 33,000 \\
RewardBench~\cite{lambert2024rewardbenchevaluatingrewardmodels} & Model Evaluation & Text & NAACL 2025 & Pair-wise & 8,108 \\
RewardBench2~\cite{malik2026rewardeval} & Model Evaluation & Text & \rev{ICLR 2026} & Pair-wise & 1,865 \\
VL-RewardBench~\cite{li2025vl} & Model Evaluation & Multimodal & CVPR 2025 & Pair-wise & 1,247 \\
\bottomrule
\end{tabular}%
}
\caption{\rev{Overview of DPO datasets.}}
\label{tab:dataset}
\end{table*}

DPO algorithm aligns LLMs using datasets that consist of preference pairs. Typically, such a dataset includes a prompt paired with two responses: a preferred (chosen) answer and a dispreferred (rejected) answer. 
Human annotators are guided to evaluate and provide preferences among multiple responses generated from the same prompt. 
These datasets are denoted as \textbf{human-labeled} datasets in this survey.
This process, however, is resource-intensive, especially when dealing with large volumes of data. 
To mitigate the high cost associated with human annotation, some datasets employ language models to synthesize preference pairs~\cite{ultrafeedback}~\cite{yu2024rlaifvaligningmllmsopensource}~\cite{zhang2024spavlcomprehensivesafetypreference}. These synthesized datasets are referred to as \textbf{AI-labeled} datasets in this survey. 
\rev{While the aforementioned datasets are constructed for training, independent benchmarks are critical to evaluate the alignment performance. In addition to general capability benchmarks, pair-wise benchmarks directly test whether the model can accurately rank different answers via its implicit reward mechanism. Therefore, we categorize the current dataset landscape into Human-Labeled Datasets, AI-Labeled Datasets, and Evaluation Benchmarks.} 

For the sake of brevity, this section highlights only a few representative datasets to illustrate their core value and construction methods. For a detailed list and more information, readers are referred to the main dataset table (Table \ref{tab:dataset}) in our paper. In the appendix, we will explore these datasets in greater detail.

\subsection{Human-Labeled Datasets}

Considered the \textbf{gold standard} for alignment, these datasets directly capture human judgements but are expensive and time-consuming to create.

\begin{itemize}
    \item \textbf{HH-RLHF (Helpful and Harmless-RLHF)~\cite{bai2022traininghelpfulharmlessassistant}:} A foundational dataset from Anthropic, \rev{pivotal} for safety alignment. It contains dialogues where human labellers chose the more ``helpful and harmless'' response between two options. Its unique ``red-teaming'' component---where labellers deliberately tried to elicit harmful responses---makes it \rev{invaluable} for training robustly safe AI.

    \item \textbf{SHP (Stanford Human Preferences)~\cite{pmlr-v162-ethayarajh22a}:} This large-scale dataset cleverly mines real-world human preferences from Reddit. It operates on the assumption that a comment with more upvotes is generally preferred over another on the same post. Covering 18 diverse subject areas, from cooking to legal advice, SHP provides a rich source of alignment signals from the wild.

    \item \textbf{OASST1 (OpenAssistant Conversations)~\cite{openassistant}:} This is a massive, crowdsourced, and multilingual dataset. It consists of conversation trees where volunteers both created and rated responses. Its tree-like structure captures complex, multi-turn interactions, making it \rev{ideal} for training sophisticated conversational agents.
\end{itemize}

% --- AI-LABELLED SECTION ---
\subsection{AI-Labeled Datasets}

To overcome the cost and scale limitations of human labelling, many datasets now use powerful models like GPT-4 to generate preference labels. This approach is often called Reinforcement Learning from AI Feedback (RLAIF).

\begin{itemize}
    \item \textbf{UltraFeedback~\cite{ultrafeedback}:} A large and diverse dataset designed for training high-quality reward models. Its process involves generating multiple responses for each prompt using various LLMs, and then using GPT-4 to provide detailed, multi-faceted feedback, including ratings for helpfulness, honesty, and truthfulness. This rich, fine-grained AI feedback offers high-quality supervision for alignment.

    \item \textbf{RLAIF-V~\cite{yu2024rlaifvaligningmllmsopensource}:} This dataset extends the AI-labelling concept to the \textbf{multimodal} domain, specifically to mitigate ``hallucinations'' in LVLMs. The method involves breaking down a model's textual response to an image into atomic facts and using an AI model to verify each fact against the visual content. This creates high-quality preference pairs that teach the model to generate visually grounded and trustworthy responses.

    \item \textbf{Nectar~\cite{starling2023}:} One of the first large-scale datasets to move beyond simple pairwise preferences. For each prompt, Nectar provides a ranked list of seven different responses, generated by various models and then ranked by GPT-4. This list-wise format offers a richer training signal than simple chosen/rejected pairs, allowing for more nuanced and precise preference alignment.
\end{itemize}

% --- Evaluation Benchmarks SECTION ---
\rev{
\subsection{Evaluation Benchmarks}
To measure the alignment quality of DPO, specialized pair-wise benchmarks are utilized. These benchmarks focus on evaluating the model's generation and the accuracy of its implicit reward ranking across diverse tasks.
\begin{itemize}
    \item \textbf{Chatbot Arena~\cite{zheng2023judgingllmasajudgemtbenchchatbot}:} A generative evaluation platform relying on crowdsourced, blind pairwise battles. It calculates Elo ratings based on human preferences, serving as an important leaderboard for real-world instruction-following capabilities.
    \item \textbf{RewardBench2~\cite{malik2026rewardeval}:} This benchmark shifts the evaluation setting from a standard two-option evaluation to a more challenging four-option evaluation (one Chosen and three Rejected). By reducing the random baseline to 25\%, it leaves more room for model improvement and evaluates challenging areas like factuality and precise instruction following.
    \item \textbf{VL-RewardBench~\cite{li2025vl}:} A benchmark specifically designed to evaluate visual-language generation reward models. It systematically probes performance in visual perception, hallucination detection, and reasoning tasks across 1,247 high-quality examples verified by human evaluation.
\end{itemize}
}

\section{Applications}
\label{ch: Applications}
Due to its efficiency and stability, DPO has become a \rev{leading} technique for aligning language and multi-modal models. This section highlights its most representative applications in \textbf{Large Language Models (LLMs)} and \textbf{Multi-modal Models}, thereby showcasing its broad impact. A more detailed list of applications can be found in the Appendix.

\subsection{Applications in Large Language Models (LLMs)}

DPO has been widely adopted by \rev{leading} industry models as a replacement for the more complex and resource-intensive RLHF pipeline.

\begin{itemize}
    \item \textbf{Alignment for State-of-the-Art Models (Llama 3~\cite{llama3herdmodels} \& Qwen2~\cite{qwen2technicalreport}):} Both Meta's \textbf{Llama 3} and Alibaba's \textbf{Qwen2} model series explicitly use DPO in their alignment stages. They confirm that, compared to traditional algorithms like PPO, DPO is not only less computationally expensive and more stable but also achieves superior performance on key capabilities like instruction following and safety. The Llama 3 team further improved upon DPO by masking special tokens in the loss calculation and adding an NLL term to enhance stability and alignment effectiveness.

    \item \textbf{Enhancing Complex Reasoning (Step-DPO~\cite{lai2024stepdpostepwisepreferenceoptimization}):} Vanilla DPO treats an entire response as a single unit for optimization, which struggles with long-chain, multi-step tasks like mathematical reasoning because it cannot pinpoint the specific step where an error occurs. To address this, methods like \textbf{Step-DPO} were introduced. It refines the optimization granularity from the entire response to individual reasoning steps, improving accuracy on complex reasoning tasks. 
    % By contrasting correct and incorrect reasoning ``steps,'' it provides the model with a more precise supervisory signal, significantly improving accuracy on complex reasoning tasks.

    \item \textbf{Mitigating Hallucinations (Factuality Alignment):} Model ``hallucinations'' (generating factually incorrect content) are a major challenge in LLM applications. Research has shown that DPO can play a key role in factuality alignment~\cite{rlhf-v,xiao2024detectingmitigatinghallucinationlarge}. By constructing preference pairs that teach the model to ``prefer'' factually accurate responses over those containing hallucinations, DPO effectively reduces the model's tendency to generate false information. This approach is often combined with techniques like Retrieval-Augmented Generation (RAG) for optimal results.
\end{itemize}

\subsection{Applications in Multi-modal Models}

The principles of DPO are equally applicable to optimizing models that generate multi-modal content like images and audio.

\begin{itemize}
    \item \rev{\textbf{Image Generation:} Adapting preference learning to continuous visual spaces requires reformulating the reverse denoising process as an optimizable policy. While Diffusion-DPO~\cite{wallace2023diffusionmodelalignmentusing} achieves this by directly optimizing text-to-image models over paired visual feedback, applying preference loss to continuous diffusion trajectories suffers from severe training instability and high annotation costs. To constrain this highly variant optimization space, SSPO~\cite{zhang2024bridging} introduces an autonomous self-sampling mechanism. By constructing preference pairs from historical checkpoints and dynamically interpolating the DPO loss with the SFT objective, SSPO strictly prevents the distribution shift of denoising trajectories while bypassing external human annotations.}

    \item \textbf{Audio-Visual Understanding and Generation:} In multi-modal understanding, DPO is used to reduce hallucinations when LVLMs describe images or videos~\cite{yu2024rlaifvaligningmllmsopensource, xiao2024detectingmitigatinghallucinationlarge}. In multi-modal generation~\cite{zhang2024speechalignaligningspeechgeneration}, it is used to fine-tune speech synthesis models for more accurate emotional expression or to match a user's preferred style better.
\end{itemize}

\subsection{Applications in Agents and Robotics}

\rev{Transitioning from static generation to dynamic interactions, preference optimization is increasingly deployed to handle sequential decision-making and continuous physical control.}

\begin{itemize}
\item \rev{\textbf{Language Agents:} Standard DPO adopts single-turn generation, limiting its deployment in sequential multi-turn environments. To address this, DMPO~\cite{dmpo} and SDPO~\cite{kong2025sdpo} adapt the preference objective for multi-turn trajectories and key interaction segments. For long-horizon tasks, HPL~\cite{hpl} resolves credit assignment mismatches by decomposing trajectories into semantic action groups via hierarchical curriculum learning. Furthermore, to prevent agents from passively responding to human errors in collaborative scenarios, FAAF~\cite{faaf} injects contextual friction during alignment, training models to proactively verify evidence and correct mistakes.}

\item \rev{\textbf{Robotics and Embodied AI:} Deploying preference optimization in physical domains bypasses dense reward engineering but requires bridging cognitive decisions with continuous control. To align cognitive planning with environmental constraints, TCPO~\cite{tcpo2025} optimizes the intermediate reasoning of embodied agents prior to physical execution. For robotic manipulation, APO~\cite{xia2025humanassisted} utilizes continuous human-assisted feedback to explicitly penalize failure-prone trajectories in Vision-Language-Action models. Expanding to continuous motion generation, SoPo~\cite{tan2025sopo} employs a semi-online framework pairing generated unpreferred motions with offline-curated trajectories to synthesize human-preferred kinematics. }
\end{itemize}

% \section{Discussion and Open Challenges}
% \label{ch:discussion}
% Numerous analytical studies and variants of DPO for large language and multi-modal models have been proposed, as detailed above.
% However, new challenges related to DPO continue to emerge.
% Therefore, several open issues remain that warrant thorough discussion and investigation. 
% In this section, we provide future research directions, including preference dataset construction, generalization of DPO, and more complex applications in the Appendix.

\section{Discussion and Open Challenges}
\label{ch:discussion}
Despite the rapid development of DPO variants, several open issues remain. In this section, we outline future research directions regarding preference datasets, DPO generalization, and broader applications.

\subsection{Preference Feedback}
Current preference datasets exhibit limitations that require future investigation:
\begin{itemize}
    \item \textbf{Annotation Scalability:} Relying heavily on proprietary models (e.g., GPT-4) or human annotators restricts dataset quality and scalability, especially as AI models begin to surpass human evaluation capabilities~\cite{lee2023rlaif,burns2023weaktostronggeneralizationelicitingstrong,zhao2024weak}.
    \item \textbf{Multi-Source Feedback:} Feedback typically originates from a single source. Effectively harnessing heterogeneous feedback from real-world environments, tools, and specialized models~\cite{instruct-code-llama, mcaleese2024llm} to improve model alignment remains largely unexplored.
    \item \textbf{Fine-Grained Feedback:} Most existing feedback is coarse-grained (instance-level). Developing efficient fine-grained (sentence/token-level) feedback~\cite{rlhf-v, fine-grained, xiao2024detectingmitigatinghallucinationlarge, lai2024stepdpostepwisepreferenceoptimization, lu2024stepcontrolleddpoleveragingstepwise} is crucial for advancing reasoning models like OpenAI o1~\cite{learningtoreason} that rely on long chain-of-thoughts.
\end{itemize}

\subsection{Generalization of DPO}
DPO often exhibits inferior generalization compared to online methods~\cite{xu2024dpo,ivison2024unpackingdpoppodisentangling}. Potential improvements include:
\begin{itemize}
    \item \textbf{Online Data Exploitation:} DPO's reliance on offline data limits its generalizability~\cite{wang2024offlinerlhfmethodsneed}. While iterative variants incorporate online data~\cite{yuan2024self,xiong2024iterativepreferencelearninghuman,pang2024iterativereasoningpreferenceoptimization}, their performance often plateaus, demanding more effective data exploitation strategies.
    \item \textbf{Nuanced Learning Objectives:} The standard DPO objective ignores the varying degrees of preference margins between pairs. Future work must enhance awareness of nuanced preference differences~\cite{meng2024simpo,xu2024contrastive,xiao2024detectingmitigatinghallucinationlarge,azar2023general} and foster intrinsic reasoning capabilities rather than just mimicking formatting.
    \item \textbf{Knowledge-Augmented Rewards:} Implicit reward models in DPO rely on static parameters, lacking real-time and domain-specific knowledge. Integrating tool- or knowledge-augmented signals~\cite{li2024toolaugmented, tian2023finetuninglanguagemodelsfactuality, lin2024flamefactualityawarealignmentlarge} remains a challenging necessity.
\end{itemize}

\subsection{More Applications}
DPO's potential extends well beyond current standard tasks:
\begin{itemize}
    \item \textbf{Mixed-Modal Models:} DPO's autoregressive formulation is highly suitable for mixed-modal models~\cite{lu2023unifiedio, Chameleon_Team_Chameleon_Mixed-Modal_Early-Fusion_2024, he2025mars,xie2024showo} that unify multi-modal understanding and generation into discrete tokens.
    \item \textbf{Advanced Reasoning:} Following the paradigm shift towards post-training computing (e.g., OpenAI o1~\cite{learningtoreason}), DPO can be adapted to reward fine-grained chain-of-thought, reinforcing correct trajectories over multiple complex reasoning steps.
    \item \textbf{Video Generation:} Adapting DPO to video diffusion models~\cite{prabhudesai2024video, li2024surveylongvideogeneration} has the potential to mitigate physical law violations, safety issues, and controllability challenges in real-world simulations.
\end{itemize}

\section{Conclusion}
\label{ch:conclusion}
In this work, we provide a review of recent advancements in DPO, a widely used lightweight preference learning method, covering aspects such as preference feedback, theoretical analyzes, variants, and applications. Additionally, we propose several future research directions, with the aim of offering insight to the research community on aligning foundational models, including LLMs, MLLMs, and beyond.

\section*{Acknowledgments}
\noindent This work was supported in part by the Ningbo Youth Science and Technology Innovation Leading Talent Program (No. 2025QL059), the "Pioneer and Leading Goose" R\&D Program of Zhejiang (No. 2025C02037), Alibaba-Zhejiang University Joint Research Institute of Frontier Technologies, the Huawei Model Merge Collaboration Project, and the Science and Technology Project of State Grid Beijing Electric Power Company (Project Title: Research on Urban Cable Network Operation Status Detection and Risk Identification Technology Based on Soft Robots and Artificial Intelligence, Project Number: 520246250003).

\bibliographystyle{IEEEtran}
\bibliography{custom}

\end{document}